\documentclass[journal]{IEEEtran}

\usepackage[mathscr]{eucal}
\usepackage[cmex10]{amsmath}
\usepackage{epsfig,epsf,psfrag}
\usepackage{amssymb,amsmath,amsthm,amsfonts,latexsym}
\usepackage{amsmath,graphicx,bm,xcolor,url}
\usepackage[caption=false]{subfig} 
\usepackage{fixltx2e}
\usepackage{array}
\usepackage{verbatim}
\usepackage{bm}
\usepackage{algorithmic, cite}
\usepackage{algorithm}
\usepackage{verbatim}
\usepackage{textcomp}
\usepackage{mathrsfs}
\usepackage{epstopdf}

\catcode`~=11 \def\UrlSpecials{\do\~{\kern -.15em\lower .7ex\hbox{~}\kern .04em}} \catcode`~=13 

\allowdisplaybreaks[3]

\newcommand{\calI}{\mathcal{I}}

\newcommand{\calL}{\mathcal{L}}

\newcommand{\calN}{\mathcal{N}}

\newcommand{\calP}{\mathcal{P}}

\newcommand{\calS}{\mathcal{S}}

\newcommand{\calU}{\mathcal{U}}

\newcommand{\ba}{\mathbf{a}}
\newcommand{\bA}{\mathbf{A}}
\newcommand{\bb}{\mathbf{b}}
\newcommand{\bB}{\mathbf{B}}
\newcommand{\bc}{\mathbf{c}}
\newcommand{\bC}{\mathbf{C}}
\newcommand{\bd}{\mathbf{d}}
\newcommand{\bD}{\mathbf{D}}
\newcommand{\be}{\mathbf{e}}

\newcommand{\bg}{\mathbf{g}}
\newcommand{\bG}{\mathbf{G}}

\newcommand{\bH}{\mathbf{H}}

\newcommand{\bI}{\mathbf{I}}

\newcommand{\bM}{\mathbf{M}}

\newcommand{\bP}{\mathbf{P}}

\newcommand{\bs}{\mathbf{s}}
\newcommand{\bS}{\mathbf{S}}

\newcommand{\bT}{\mathbf{T}}
\newcommand{\bu}{\mathbf{u}}
\newcommand{\bU}{\mathbf{U}}
\newcommand{\bv}{\mathbf{v}}
\newcommand{\bV}{\mathbf{V}}

\newcommand{\bx}{\mathbf{x}}
\newcommand{\bX}{\mathbf{X}}

\newcommand{\bY}{\mathbf{Y}}

\newcommand{\dU}{\dot{\mathbf{U}}}
\newcommand{\dD}{\dot{\mathbf{D}}}

\DeclareMathAlphabet{\mathbsf}{OT1}{cmss}{bx}{n}
\DeclareMathAlphabet{\mathssf}{OT1}{cmss}{m}{sl}

\DeclareSymbolFont{bsfletters}{OT1}{cmss}{bx}{n}  
\DeclareSymbolFont{ssfletters}{OT1}{cmss}{m}{n}
\DeclareMathSymbol{\bsfGamma}{0}{bsfletters}{'000}
\DeclareMathSymbol{\ssfGamma}{0}{ssfletters}{'000}
\DeclareMathSymbol{\bsfDelta}{0}{bsfletters}{'001}
\DeclareMathSymbol{\ssfDelta}{0}{ssfletters}{'001}
\DeclareMathSymbol{\bsfTheta}{0}{bsfletters}{'002}
\DeclareMathSymbol{\ssfTheta}{0}{ssfletters}{'002}
\DeclareMathSymbol{\bsfLambda}{0}{bsfletters}{'003}
\DeclareMathSymbol{\ssfLambda}{0}{ssfletters}{'003}
\DeclareMathSymbol{\bsfXi}{0}{bsfletters}{'004}
\DeclareMathSymbol{\ssfXi}{0}{ssfletters}{'004}
\DeclareMathSymbol{\bsfPi}{0}{bsfletters}{'005}
\DeclareMathSymbol{\ssfPi}{0}{ssfletters}{'005}
\DeclareMathSymbol{\bsfSigma}{0}{bsfletters}{'006}
\DeclareMathSymbol{\ssfSigma}{0}{ssfletters}{'006}
\DeclareMathSymbol{\bsfUpsilon}{0}{bsfletters}{'007}
\DeclareMathSymbol{\ssfUpsilon}{0}{ssfletters}{'007}
\DeclareMathSymbol{\bsfPhi}{0}{bsfletters}{'010}
\DeclareMathSymbol{\ssfPhi}{0}{ssfletters}{'010}
\DeclareMathSymbol{\bsfPsi}{0}{bsfletters}{'011}
\DeclareMathSymbol{\ssfPsi}{0}{ssfletters}{'011}
\DeclareMathSymbol{\bsfOmega}{0}{bsfletters}{'012}
\DeclareMathSymbol{\ssfOmega}{0}{ssfletters}{'012}

\DeclareMathOperator*{\argmax}{arg\,max}

\newtheorem{theorem}{Theorem} 
\newtheorem{lemma}[theorem]{Lemma}

\newtheorem{definition}{Definition}

\newtheorem{remark}{Remark}

\newtheorem{assumption}{Assumption}
\newtheorem{data model}{Data Model}
\newtheorem{requirement}{Requirement}

\newcommand{\qednew}{\nobreak \ifvmode \relax \else
      \ifdim\lastskip<1.5em \hskip-\lastskip
      \hskip1.5em plus0em minus0.5em \fi \nobreak
      \vrule height0.75em width0.5em depth0.25em\fi}

\begin{document}
\title{Provable Clustering of a Union of Linear Manifolds Using Optimal
 Directions}

\author{\textbf{Mostafa~Rahmani}\\
Amazon  
\thanks{Email: mostrahm@amazon.com}
}

\markboth{}%
{Shell \MakeLowercase{\textit{et al.}}: Bare Demo of IEEEtran.cls for Journals}
\maketitle

\begin{abstract}
This paper focuses on the Matrix Factorization based Clustering (MFC) method  which is one of the few closed-form algorithms for the subspace clustering problem. Despite being simple, closed-form, and computation-efficient, MFC can outperform the other sophisticated subspace clustering methods in many challenging scenarios. We reveal the connection between MFC and the Innovation Pursuit (iPursuit) algorithm which was shown to be able to outperform the other spectral clustering based methods with a notable margin especially when the span of clusters are close. A novel theoretical study is presented which sheds light on the key performance factors of both algorithms (MFC/iPursuit) and  it is shown that both algorithms can be robust to notable intersections between the span of  clusters. Importantly, in contrast to the theoretical guarantees of other algorithms which emphasized on the distance between the subspaces as the key performance factor and without making the innovation assumption, it is shown that the performance of MFC/iPursuit mainly depends on the distance between the innovative components of the clusters.
\end{abstract}

\section{Introduction}
When data points lie in a single linear manifold, conventional techniques such as Principal Component Analysis (PCA)  can be efficiently used to find the underlying low-dimensional structure~\cite{zhang2014novel,lerman2015robust}. 
However, in many applications, the data points may be originating from multiple independent sources and a union of  manifolds can better model the data \cite{vidal2011subspace}.
The subspace clustering problem is defined on how to learn these low dimensional  manifolds when they are linear subspaces \cite{heckel2013robust,elhamifar2013sparse,DBLduallll,rahmani2017innovation,peng2016deep,lu2013correlation,feng2014robust,patel2013latent,wang2013provable,lu2013correlation,wang2016noisy,you2016scalable,ji2017deep,zhang2018cappronet,klys2018learning,peng2016deep, menon2020subspace,jiang2018nonconvex,lipor2021subspace} in a completely unsupervised way.



\noindent
\textbf{Summary of contributions:} 
This paper focuses on analyzing two  subspace clustering algorithms:  Matrix Factorization based Clustering (MFC) and Innovation Pursuit (iPursuit).
First we reveal the underlying connection between them and the presented analysis shows why they can notably outperform other spectral clustering based methods in the challenging scenarios. The main contributions of this work can be summarized as follows. 

\noindent
$\bullet$ It is shown that iPursuit is equivalent to MFC  if we alter its $\ell_1$-norm based cost function  into a quadratic cost function and importantly, all the presented theoretical results are applicable to both algorithms. 

\noindent
$\bullet$ 
To the best of our knowledge, this paper presents the first comprehensive analysis of MFC/iPursuit algorithms and the presented analysis is not based on the restrictive innovation assumption used in \cite{rahmani2017innovation,rahmani2017innovationJ,li2021provable}. 
The MFC/iPursuit algorithms are analyzed and we establish  deterministic and probabilistic sufficient conditions which guarantee that the computed adjacency matrix by MFC/iPursuit satisfies a defined quality requirement. Importantly, it is shown that in contrast to most of other clustering algorithms whose performance depend on the distance between the subspaces, the performance of MFC/iPursuit mainly depends on the distance between the innovative components of the clusters. Accordingly, even if the span of clusters intersect heavily, MFC/iPursuit can still provably satisfy the performance requirement.


\subsection{Notation and Definitions}
Given a matrix $\bA$, $\| \bA \|$ denotes its spectral norm, $\| \bA \|_F$ denotes its Frobenius norm, and $\| \bA \|_{p,1} = \sum_{i} \| \ba_i \|_p$ where $\ba_i$ denotes the $i^{th}$ column of $\bA$ and $\ba^i$ denotes the $i^{th}$ row of $\bA$. For a vector $\ba$, $\| \ba \|_p$ denotes its $\ell_p$-norm, $\ba(i)$ denotes its $i^{\text{th}}$ element, and $\ba[i:k]$  contains the elements of $\ba$ whose indexes are from $i$ to $k$.   The elements of matrix $\bY = |\bX|$ are equal to the absolute value of the elements of matrix $\bX$.
The subspace $\calU^{\perp}$ is the complement of $\calU$. $\mathbb{S}^{M_1 - 1}$ indicates the unit $\ell_2$-norm sphere in $\mathbb{R}^{M_1}$. It is assumed that data matrix $\bD \in \mathbb{R}^{M_1 \times M_2}$ can be represented as $\bD = \bU \Sigma \bV^T$ where $\bU \in \mathbb{R}^{M_1 \times r_d}$ is the matrix of left singular vectors, the diagonal matrix $\Sigma \in \mathbb{R}^{r_d \times r_d}$ contains the non-zero singular values,  the columns of $\bV \in \mathbb{R}^{  M_2 \times r_d}$ are equal to the right singular vectors,  $r_d$ is the rank of $\bD$, $M_2$ is the number of data points, and $M_1$ is the dimension of ambient space. The subspace $\calS = \oplus_{i=1}^m \calS_i$ is equal to the direct sum of subspaces $\{\calS_i \}_{i=1}^m$ and $\text{dim}(\calS)$ denotes the dimension of $\calS$.  Two adjacency matrices $\bA \in \mathbb{R}^{M_2\times M_2}$ and $\bB\in \mathbb{R}^{M_2\times M_2}$ are said to be equivalent when $\frac{\ba_i}{\| \ba_i\|_1} = \frac{\bb_i}{\| \bb_i\|_1}$ holds for all $1\le i \le M_2$. RHS means right hand side and LHS means left hand side. 

\noindent
\textbf{Distance between subspaces:} Suppose $\bU_1 \in \mathbb{R}^{M_1 \times r}$ and $\bU_2 \in \mathbb{R}^{M_1 \times r}$ are orthonormal bases for r-dimensional subspaces $\calS_1$ and $\calS_2$, respectively. Two different notions are used to express the affinity between two subspaces. One measure is $\| \bU_1^T \bU_2 \|$. However, $\| \bU_1^T \bU_2 \|$ is always equal to 1 when $\text{dim}(\calS_1 \cap \calS_2) >0$. The other measure of affinity between two subspaces is $$\| \bU_1^T \bU_2 \|_{\sigma} = \sqrt{\frac{\sum_{i=1}^r \cos^2 \theta_i}{r}}$$ where $\{\theta_i \}_{i=1}^r$ are the principal angles between $\calS_1$ and $\calS_2$ \cite{soltanolkotabi2012geometric}. Note that $\|\bU_1^T \bU_2 \|_{\sigma} =1$ only when $\calS_1 = \calS_2$. 

\subsection{Data Model}
Data Model 1 provides the details of the presumed model along with  definition of the used symbols. To simplify the exposition and the analysis, it is assumed that the dimension of subspaces are equal, the number of data points in different clusters are equal, and a subspace $\calS$ is used to define the intersection between the span of clusters. 

\begin{data model}
The data matrix $\bD \in \mathbb{R}^{M_1 \times M_2}$ can be written as $$\bD = [\bD_1 \:,\: \bD_2 \:, ...\:, \bD_m] \bT$$
where $\bT\in \mathbb{R}^{M_2 \times M_2}$
is an unknown permutation matrix. We define $\calS_i$ as the column space of $\bD_i$ and  $\calS_i \not\subset \calS_j$ and $\calS_j \not\subset \calS_i$ for any $i \neq j$. 
The dimension of all subspaces is equal to $r$ and there are $n$ data points in each cluster, i.e., $\bD_i \in \mathbb{R}^{M_1 \times n}$. The dimension of the intersection between subspaces is equal to s, i.e., $\text{dim} \left(\cap_{i=1}^m \calS_i\right) = s$ and we define subspace $\calS = \cap_{i=1}^m \calS_i$. In addition, $\calS_i \cap \calS_j=\calS$ for all $i \neq j$. The orthonormal matrix $\bU_i \in \mathbb{R}^{M_1 \times r}$ is a basis for $\calS_i$ and it can be written as $\bU_i = [\bS \:,\: \dot{\bU}_i]$ where orthonormal matrix $\bS \in \mathbb{R}^{M_1 \times s}$
is a basis for $\calS = \cap_{i=1}^m \calS_i$ and $\dU_i \in \mathbb{R}^{M \times (r-s)}$ is a basis for $\calS_i \cap \calS^{\perp}$. The orthonormal matrix $\dU_i $ represents the component of $\calS_i$ which does not lie in $\calS$ and we call $$\dot{\calS}_i = \text{span}(\dU_i) = \calS_i \cap \calS^{\perp}$$ the innovative component of $\calS_i$.
Each data point $\bd_i$ which lies in $\calS_{k_i}$ can be represented as 
\begin{eqnarray}
\bd_i = \bS \alpha_i + \dU_{k_i} \beta_i \:,
\label{eq:alpha_beta}
\end{eqnarray}
where $\alpha_i \in \mathbb{R}^{s}$ and $\beta_i \in \mathbb{R}^{r-s}$. 
\end{data model}

In order to  represent the association of  each data point to its corresponding cluster, we define index $k_i$ such that $\bd_i \in \calS_{k_i}$. Matrix $\bD_{-k}$ includes all the columns of $\bD$ except the ones which lie in $\calS_k$. Matrices $\dD_j$ and $\bar{\bD}_j$ are defined as  $\dD_j = \dU_j^T \bD_j$ and $\bar{\bD}_j = \bS^T \bD_j$.

\begin{algorithm}
\caption{Data Clustering Using iPursuit }
{
\textbf{Input.} The input is data matrix $\bD \in \mathbb{R}^{M_1 \times M_2}$.

\smallbreak
\textbf{1. Project data points on $\mathbb{S}^{M - 1}$.}  Set $\bd_i$ equal to $\bd_i / \| \bd_i \|_2$ for all $1 \le i \le M_2$.

\smallbreak
\textbf{2. Direction search.}
Define $\bC^{*} \in \mathbb{R}^{M_1 \times M_2}$   as  optimal point of
$$
\underset{ \bC}{\min} \: \:  \| \bC^T \bD \|_{1} \: \: \text{subject to} \: \: \text{diag}(\bC^T \bD) = \textbf{1} \:.
\label{opt:kolli}
$$
\smallbreak

\textbf{3.} Define adjacency matrix $\bA= \big| \bC^T \bD \big|$.

\textbf{4.} Apply  graph preprocessing steps (e.g., sparsifying adjacency matrix $\bA$ via keeping few dominant non-zero elements of each row).

\textbf{5.} Apply spectral clustering  to $\bA + \bA^T$.

\textbf{Output:} The identified clusters.

 }
\end{algorithm}

\section{Related Work}
Numerous approaches for subspace clustering were proposed in prior work including statistical-based approaches \cite{yang2006robust,stat1,stat2,rnc1}, spectral clustering based methods \cite{elhamifar2013sparse,liu2013robust}, the algebraic-geometric approach \cite{vidal2005generalized}, and iterative methods \cite{bradley2000k}.   Much of the recent research work on subspace clustering is focused on spectral clustering \cite{von2007tutorial} based methods \cite{dyer2013greedy,gao2015multi,elhamifar2013sparse,heckel2013robust,liu2013robust,rahmani2017subspace,soltanolkotabi2012geometric,wang2013provable,chen2009spectral,park2014greedy}.

The spectral clustering based algorithms are composed of two main steps and they only differ in the first step. First, an adjacency matrix is constructed via finding a neighborhood set for each data point and in the second step, the spectral graph clustering algorithm \cite{von2007tutorial} is applied to the learned adjacency matrix. 
For instance, Sparse Subspace Clustering (SSC)~\cite{elhamifar2013sparse} uses $\ell_1$-minimization to construct a  sparse adjacency matrix, Low-Rank Representation (LRR) \cite{liu2013robust} uses nuclear norm minimization to find the adjacency matrix, and 
the Thresholding based Subspace Clustering (TSC) method \cite{heckel2013robust} simply uses the inner-product between the data points to construct the adjacency matrix. In contrast to TSC which uses  inner-product between the data points to construct the adjacency matrix,  iPursuit    \cite{rahmani2017innovation,rahmani2017subspace} utilized the directions of  innovation to measure the similarity between the data points. 
The Matrix Factorization based Clustering (MFC) method \cite{kanatani2001motion,costeira1998multibody,boult1991factorization}  is a closed-from spectral clustering based method which utilizes the right singular vectors of the data to construct the adjacency matrix.

\begin{algorithm}
\caption{Matrix Factorization based Clustering (MFC)}
{
\textbf{Input.} The input is data matrix $\bD \in \mathbb{R}^{M_1 \times M_2}$.

\smallbreak
\textbf{1. Project data points on $\mathbb{S}^{M - 1}$.}  Set $\bd_i$ equal to $\bd_i / \| \bd_i \|_2$ for all $1 \le i \le M_2$.

\smallbreak
\textbf{2. SVD:} Compute $\bD = \bU \mathbf{\Sigma}\bV^T$ where the columns of $\bV \in \mathbb{R}^{  M_2 \times r_d}$ are equal to the right singular vectors.

\textbf{3.} Define $\bA = \big| \bV \bV^T|$.

\textbf{4.} Similar to Step 4 in Algorithm 1. 

\textbf{5.} Similar to Step 5 in Algorithm 1. 

\textbf{Output:} The identified clusters.

 }
\end{algorithm}

\subsection{A Brief Overview of iPursuit (Algorithm 1)}
Suppose that data matrix $\bD$ follows Data Model 1. If the span of clusters satisfy Assumption \ref{asm:innnov}, then we say that  Innovation Assumption holds. 
\begin{assumption}
For each subspace $\calS_i$, we have $\calS_{i} \notin \underset{k\neq i}{\oplus} \calS_k$. 
\label{asm:innnov}
\end{assumption}
Define orthonormal matrix $\bP_i$ such that the column-space of $\bP$ is equal to $\calP_i = \underset{k\neq i}{\oplus} \calS_k$.
If the innovation assumption holds, then the rank of $(\bI - \bP_i \bP_i^T) \bU_i$ is greater than zero and we define $\Vec{\calS}_i$ as the column-space of $(\bI - \bP_i \bP_i^T) \bU_i$. The 
geometrical idea behind
 iPursuit  is that if we can find a direction in $\Vec{\calS}_i$, it is orthogonal to all the clusters except $\calS_i$ and this fact can be used to distinguish $\calS_i$ from the rest of clusters. Specifically, in order to find a direction in $\Vec{\calS}_{k_i}$ corresponding to each $\bd_i$, \cite{rahmani2017innovation,rahmani2017subspace} proposed to find this direction (dubbed the direction of innovation corresponding to $\bd_i$) as the optimal point of
\begin{eqnarray}
\underset{ \bc}{\min} \: \:  \| \bc^T \bD \|_1 \quad \text{subject to} \qquad \bc^T \bd_i = 1 \:.
\label{eq:intro_ip_1}
\end{eqnarray}
The motivation behind the design of (\ref{eq:intro_ip_1}) was that
 the direction of innovation corresponding to $\bd_i$  can be computed via looking for a vector which is orthogonal to the maximum number of data points. 
Although the innovation assumption was used to design iPursuit, in \cite{rahmani2017subspace,rahmani2017innovation} it was numerically shown that it is not essential in the performance of iPursuit.

The authors of \cite{rahmani2017subspace,rahmani2017innovation} presented an analysis of (\ref{eq:intro_ip_1}) which is limited to a two cluster scenario and it was based on the Innovation Assumption to prove that the optimal point of (\ref{eq:intro_ip_1}) lies in $\Vec{\calS}_{k_i}$. In  contrast, the presented theoretical study (a) does not require
the innovation assumption, (b)  guarantees a completely different requirement, (c)  is the first thorough analysis of
MFC, (d) reveals the connection between iPursuit and MFC, and importantly (e) it  shows the importance
of the incoherence between the innovative components.

\section{Analyzing A Spectral Clustering based   Method}
The difference between different spectral clustering based  algorithms is in the way that they compute the adjacency matrix.  
Accordingly, we should define proper metrics using which we could determine how accurate/useful is the estimated adjacency matrix. 
The authors of \cite{soltanolkotabi2012geometric} used the number of false connections (any non-zero connection between two nodes/data-points while they belong to different clusters) as a metric to assess the estimated adjacency matrix. However, the graph clustering algorithms such as spectral clustering can yield an exact clustering of the data even if there are a significant amount of false connections in the estimated adjacency matrix provided that the estimated weights on the true connections are sufficiently stronger than the weights of the false connections. 
Therefore, in this paper,
we use the following criteria to assess the quality of a adjacency matrix and we analyze the subspace clustering algorithms to reveal if/how they satisfy Requirement 1. 

\begin{requirement}
Suppose $\bA \in \mathbb{R}^{M_2 \times M_2}$ is 
the estimated adjacency matrix.
We require all the columns of $\bA$ to satisfy
$$
 \frac{\kappa}{m-1} \| {\ba_i}_{\calI_i^{\perp}} \|_p^p < \| {\ba_i}_{\calI_i} \|_p^p  \:,  
$$
 where  $\calI_i = \{j \:\:\: | \:\: \: k_i = k_j \}$,   $\calI_i^{\perp} = \{j \:\:\: | \:\: \: k_i \neq k_j \}$,  ${k_i} = \argmax_{j} \| \bU_j^T \bd_i \|_2$, and ${\ba_i}_{\calI_i}$ contains the elements of $\ba_i$ whose indexes are in $\calI_i$.  \label{asm:sufficeint}
\end{requirement}

\noindent
The parameter $\kappa$ is chosen greater than 1 and it determines how well the  adjacency matrix represents the clustering structure of the data. Evidently, the higher is $\kappa$, the more challenging it is for a subspace clustering algorithm to satisfy Requirement \ref{asm:sufficeint}. In the following sections, we discuss the role of parameter $p$ and we analyze MFC/iPursuit such that they satisfy Requirement 1 with $p=1$/$p=2$.  

\begin{remark}
Even if $\bA$  satisfies Requirement 1 with a large $\kappa$, it does not necessarily mean that Spectral Clustering yields exact clustering. Similarly, proving that $\bA$ does not contain any false connection (as  in \cite{soltanolkotabi2012geometric}) also does not guarantee  exact clustering. However, these measures are useful to assess how clear the estimated $\bA$ represents the clustering structure. In addition, although Requirement 1 does not guarantee exact clustering by the spectral clustering step, it is very similar to the sufficient condition stated in \cite{ling2020certifying} to guarantee that the spectral clustering algorithm yields the exact clustering. 
Specifically, \cite{ling2020certifying} proves that if 
$$
\max_{i} \| {\ba_i}_{\calI_i^{\perp}} \|_1 < \frac{\min_{k} \gamma_{2}(\calL(A_k))}{4}\:,
$$
then the spectral clustering algorithm studied in \cite{ling2020certifying} yields  an exact clustering where $\gamma_{2}(\calL(A_k))$ is the second smallest eigenvalue of graph Laplacian w.r.t. the $k^{th}$ cluster and $\bA_k \in \mathbb{R}^{n \times n}$.

\end{remark}

\section{Theoretical Studies}
This section focuses on analyzing  MFC/iPursuit  and revealing the key factors in its performance. First, we discuss the  underlying  connection between iPursuit and MFC
and this interesting connection is utilized to analyze both algorithms using similar techniques.  
 In the presented results, we  utilize the parameters defined in the following definition.
 \begin{definition}
Suppose $\bD$ follows Data Model 1. We define
$\Delta_{\min} = \min_{j} \{ \underset{\|\bu\| = 1 \atop \bu \in \calS_{j}}{\inf} \:\| \bu^T \bD_{j} \|_p^p \}_{j=1}^{m}$, $
 \dot{\Delta}_{\max} =  \max_{j} \{ \underset{\|\bu\| = 1 \atop \bu \in \mathbb{R}^{r-s}}{\sup} \|  \bu^T \dD_j \|_p^p \}_{j=1}^m$, 
$ \bar{\Delta}_{\max} = \max_{j} \{ \underset{\|\bu\| = 1 \atop \bu \in \mathbb{R}^{s}}{\sup} \|  \bu^T \bar{\bD}_j \|_p^p \}_{i=1}^m$ ,and $ \phi = \max_{j \neq t} \|\dU_t^T \dU_j \|$. 
In addition, when $y > x$, we define ${\sigma_{l}}(\frac{x}{y}, \delta) = \frac{x - 2\sqrt{x \log\frac{2M_2}{\delta} }}{y + 2\sqrt{ (y-x) \log \frac{2M_2}{\delta}} + 2 \log\frac{2M_2}{\delta} -  2\sqrt{x \log\frac{2M_2}{\delta} }} $ and ${\sigma_{u}}(\frac{x}{y}, \delta) = \frac{x + 2\sqrt{x \log\frac{2M_2}{\delta} } + 2 \log\frac{2M_2}{\delta}}{y + 2\sqrt{ x \log \frac{2M_2}{\delta}} + 2 \log\frac{2M_2}{\delta} -  2\sqrt{(y-x) \log\frac{2M_2}{\delta} }} $.
\end{definition}

The parameters $\Delta_{\min}$, $\dot{\Delta}_{\max}$, and $\bar{\Delta}_{\max}$ are similar to permeance statistic \cite{lerman2015robust} which indicates how well the data points are distributed inside the subspaces. 
For instance, when the columns of $\bD_i$ in $\calS_i$ 
 are concentrated around a direction,
 the value of 
 $\underset{\|\bu\| = 1 \atop \bu \in \calS_{i}}{\inf} \:\| \bu^T \bD_{i} \|_p^p$ is small in comparison to when the data points are uniformly distributed in $\calS_i$. 
Although the permeance statistic appears in the presented results, it does not necessarily mean that
 iPursuit  and MFC  require a uniform distribution of data pints inside the subspaces and the reason that it appears  is that the sufficient conditions guarantee the performance under the worst case scenarios.  The parameter $\phi$ indicates how close the innovative components $\{\dot{\calS}_i \}_{i=1}^m$ are to each other. 
  \begin{remark}
 It is important to note that $\phi$ only measures the affinity between the innovative components  $\{\dot{\calS}_i \}_{i=1}^m$. In other word, even if two subspaces $\calS_i$ and $\calS_j$ heavily intersect such that $\|\bU_i^T \bU_j \|_{\sigma}$ is nearly equal to 1, $\|\dU_i^T \dU_j \|$   could be small if the innovative components are incoherent with each other. In the following results, it is shown that in contrast to most of subspace segmentation methods whose performance depend on $\max_{j \neq t} \|\bU_t^T \bU_j \|_{\sigma}$, the performance of iPursuit and MFC mainly depends on the distance between the innovative components.
 \end{remark}

\subsection{The  Connection Between iPursuit and MFC}
The cost function of iPursuit (\ref{eq:intro_ip_1}) encourages the optimal direction $\bc_i^{*}$ to be orthogonal to the maximum number of data points.
If the innovation assumption (Assumption \ref{asm:innnov}) holds and $\bc_i^{*} \in \vec{\calS_{k_i}}$ for all the data points, then $\bA = | \bD^T \bC^{*} |$ 
does not include any false connection. 
 However, in practice the innovation assumption is not essential and $\bA = |\bD^T \bC^{*}|$ can  yield an accurate clustering of the data  even if $| \bD^T \bC^{*}|$ is not a sparse matrix \cite{rahmani2017subspace,ling2020certifying}. 
 A direct conclusion is that it may not be essential to employ $\ell_1$-norm in the cost function of (\ref{eq:intro_ip_1}). Accordingly, in this section, we investigate an iPursuit algorithm whose $i^{th}$ optimal direction is obtained as the optimal point of
\begin{eqnarray}
\underset{ \bc}{\min} \: \:  \| \bc^T \bD \|_2 \quad \text{subject to} \qquad \bc^T \bd_i = 1 \:.
\label{eq:ell2}
\end{eqnarray}
The following lemma shows that the  iPursuit algorithm which employs $\ell_2$-norm to compute the optimal directions is equivalent to MFC. 
\begin{lemma}
Define $\bC^{*}$ as the optimal point of
$$
\underset{ \bC}{\min} \: \:  \| \bD^T \bC \|_{2,1} \quad \text{subject to} \qquad \text{diag}(\bC^T \bD) = \textbf{1} \:,
$$
and define $\bA = \left| \bD^T \bC^{*}  \right|$. Then
$
\bA(i,j) = \frac{|{\bv^i}^T \bv^j|}{\| \bv^i \|_2^2} \:.
$
\label{lm:sade}
\end{lemma}
\noindent
Lemma \ref{lm:sade} shows that  iPursuit is equivalent to MFC when $\ell_2$-norm is employed to compute the optimal vectors. 
We leverage this connection between MFC and iPursuit to provide an analysis which is applicable to both algorithms. In the following theoretical results, $p$ appears as a parameter in the sufficient conditions. If $p=1$,  the sufficient condition  corresponds to iPursuit and if $p=2$, then the sufficient condition corresponds to MFC.


\subsection{An Analysis for MFC and iPursuit}
\label{sec:analysis1}
 The following theorem provides a sufficient condition to guarantee that Requirement 1 is satisfied. The presented results are applicable to both iPursuit and MFC since it is assumed that $\bA = |\bD^{T} \bC^{*}|$ where 
  $\bC^{*}$ is obtained via solving 
\begin{eqnarray}
\underset{ \bC}{\min} \: \:  \| \bD^T \bC \|_{p,1} \quad \text{subject to} \qquad \text{diag}(\bC^T \bD) = \textbf{1} \:.
\label{eq:main_with_p}
\end{eqnarray}

\begin{theorem}
Suppose that $\bD$ follows Data Model 1 and  $\bA = |\bD^{T} \bC^{*}|$ where $\bC^{*}$ is the optimal point of (\ref{eq:main_with_p}). If
 \begin{eqnarray}
  \begin{aligned}
 & \min_{i} \frac{\| \beta_i \|_2^p}{\| \bd_i \|_2^p} \: \Delta_{\min} \ge \\
 & {\dot{\Delta}_{\max} \: \kappa \left( \frac{1}{\kappa + (m-1)} +  \frac{m-1}{\kappa + (m-1)} \phi^p \right) } \:,
  \end{aligned}
\label{eq:sfip}
\end{eqnarray}
then $\bA$ satisfies  Requirement 1. 
\label{thm: 1}
\end{theorem}

In  contrast to former theoretical results which require $\max_{j \neq t} \|\bU_t^T \bU_j \|_{\sigma}$ to be sufficiently small, the presented guarantee is concerned with $\max_{j \neq t} \|\dU_t^T \dU_j \|$ and note that $\max_{j \neq t} \|\dU_t^T \dU_j \|$ can stay small even if the subspaces have a high dimension of intersection   (i.e., $\|\bU_t^T \bU_j \|_{\sigma}$ is nearly equal to 1). When $m$, the number of clusters, is large, the sufficient condition can be roughly simplified into 
$
\phi^p \le \frac{\Delta_{\min} }{\kappa\: \dot{\Delta}_{\max}} \: \min_{i} \frac{\| \beta_i \|_2^p}{\| \bd_i \|_2^p}\:,
$
which means that the higher is the dimension of intersection, the more distanced the innovative components should be. The sufficient condition requires all the data points to have a sufficiently strong projection on the innovative component. 

\subsection{Probabilistic Guarantees}
In this section, we simplify the result presented in Theorem \ref{thm: 1} in two steps.
First, we presume a random model for the distribution of the data points
and in the second step, we consider a random model for the generation of the subspaces. 
We start with the first step as follows. 
\begin{assumption}
Each matrix $\bD_i \in \mathbb{R}^{M_1 \times n}$ is generated as $\bD_i = \bU_i \bG_i$ where the elements of $\bG_i \in \mathbb{R}^{r \times n}$ are sampled independently from $\calN(0, \frac{1}{\sqrt{r}})$. 
\label{asm:data_random}
\end{assumption}

\noindent
Assumption \ref{asm:data_random} ensures
that the distribution of $\frac{\bd_i}{\| \bd_i \|_2}$ is 
uniformly at random on $\mathbb{S}^{M_1 - 1} \cap	\calS_{k_i}$.
Note that $\mathbb{E}[\| \bd_i \|_2^2] = 1$ and in the following theorems, we do not normalize the $\ell_2$-norm of the data points to make the analysis easier.  In this section, we derive the guarantees for $p=2$ and similar guarantees for $p=1$ can be established. 

\begin{theorem}
Suppose $\bD$ follows Data Model 1, matrices $\{ \bD_i \}_{i=1}^{m}$ are generated as in Assumption \ref{asm:data_random}, and adjacency matrix $\bA$ is computed as in Theorem \ref{thm: 1} with $p=2$. If
 \begin{eqnarray}
  \begin{aligned}
& (\frac{n}{r} - {\eta_{\delta}}_r ) \sigma_{l}\left( \frac{r-s}{r}, \delta \right) \ge \\
 &\quad \quad\kappa (\frac{1}{\kappa+ m-1} + \phi^2 \frac{m-1}{\kappa +m -1}) \left( \frac{n}{r} + \frac{r-s}{r}{\eta_{\delta}}_{r-s} \right)
   \end{aligned}
 \end{eqnarray}
where ${\eta_{\delta}}_{x} = \max ( \frac{4 {z_{\delta}}_x }{3} \log \frac{2\:x\:m}{\delta} , \sqrt{4 \frac{n(x+3)}{x^2} \log \frac{2 x m}{\delta}} )$ and \textcolor{black}{${z_{\delta}}_x = 1 + 2 \sqrt{\frac{1}{x} \log\frac{2n m}{\delta} } + \frac{2}{x} \log\frac{2 n m}{\delta}$}, then  Requirement \ref{asm:sufficeint} with $p=2$ is satisfied with probability at least $1 - 5\delta$.
\label{thm:random_point}
\end{theorem}

Theorem \ref{thm:random_point} reveals several interesting points about the requirements of the algorithms. First it confirms our intuition about the relation between the dimension of subspaces and the required number of data points. The sufficient condition states that $n/r$ should be sufficiently large to ensure that  Requirement 1 is satisfied. When $n/r$ is sufficiently large, then $(\frac{n}{r} - {\eta_{\delta}}_r)$ is nearly equal to $n/r$. Therefore, when $m$ is large, the sufficient condition roughly states that $\phi^2$ should be sufficiently smaller than $\frac{1}{\kappa}\frac{r-s}{r}$. In other word, Theorem \ref{thm:random_point} clearly indicates that the higher is the dimension of intersection, the more separable their innovative components should be. Next, we  further simplify  the sufficient condition  via assuming a random model for the distribution of subspaces.  
\begin{theorem}
Suppose $\bD$ and $\bA$ are generated as in Theorem \ref{thm:random_point} and    $\{\dot{\calS}_i \}_{i=1}^m$ and $\calS$ are chosen independently and uniformly at random. If
 \begin{eqnarray}
   \begin{aligned}
& \left(\frac{n}{r} - {\eta_{\delta}}_r \right) \sigma_{l}\left( \frac{r-s}{r}, \delta \right) \ge \\
& \kappa \left( \frac{n}{r} + \frac{r-s}{r}{\eta_{\delta}}_{r-s} \right)  \Big(\frac{1}{\kappa+ m-1} + \\
& \quad\quad\quad\quad\quad\frac{c_{{\delta}} (r-s)^2 }{M_1} \frac{m-1}{ \kappa + m -1} \Big) 
   \end{aligned}
 \end{eqnarray}
then Requirement \ref{asm:sufficeint} is satisfied with probability at least $1 - 6\delta$, where $c_{\delta} = 3 \max \left(1 ,  \sqrt{\frac{8  M_1 \pi }{(M_1 - 1)(r-s)}} , \sqrt{\frac{16 M_1 \log \frac{m r}{\delta} }{(M_1 - 1)(r-s) }} \right)$.
\label{thm:full_random}
\end{theorem}
If we simplify the sufficient condition, Theorem \ref{thm:full_random} roughly states that  $M_1$ should be sufficiently larger than $\kappa r (r-s) \sqrt{\log m}$. The main reason is  that the subspaces and their innovative components are generated uniformly at random and the higher is the dimension of the ambient space, the less coherent they are in expectation. 

\begin{remark}
The main purpose of the presented analysis is to demonstrate the key performance factors of the MFC/iPursuit algorithms and to show why they are notably robust to the strong intersection between the span of clusters. If we want to go further and use the theoretical results to compare MFC/iPursuit against the other subspace clustering algorithms, we need to analyze the other methods using the utilized criteria (Requirement 1). Although it goes beyond the scope of this paper, Section \ref{sec:compTSC} presents a full analysis of the TSC algorithm based on Requirement 1 to show why MFC can strongly outperform TSC while their computation complexities are not much different. 
\end{remark}

\subsection{With the Innovation Assumption}
\label{sec:with_innov}
The innovation assumption (Assumption \ref{asm:innnov}) is not essential in the performance of MFC/iPursuit and we did not use it in any of the presented studies. However, the innovation assumption can be utilized to establish \textcolor{black}{stronger} guarantees. 
In this section,  two theorems are presented whose only
difference with Theorem \ref{thm: 1} and Theorem \ref{thm:full_random} is that they assume that Assumption \ref{asm:with_inv} (stated bellow) holds. 
\begin{assumption}
It is assumed that $\bD$ follows Data Model 1 and  $\text{dim}(\dot{\calS_i} \cap	 \calP_i) = 0$ where $\calP_i =  \oplus_{k\neq i} \calS_i$.
\label{asm:with_inv}
\end{assumption}
Assumption \ref{asm:with_inv}
ensures that each innovative component  $\dot{\calS}_i$ is independent from the direct sum of all the other subspaces. The following theorem presumes that Assumption \ref{asm:with_inv} holds.
\begin{theorem}
\noindent
Suppose $\bD$ follows Assumption \ref{asm:with_inv},  define $\Vec{\calS_i}$ as the column space of $(\bI - \bP_i \bP_i^T) \bU_i$, define $\vec{\bU}_i$ as a basis for $\Vec{\calS_i}$, and assume $\bA = |\bD^{T} \bC^{*}|$ where $\bC^{*}$ is the optimal point of (\ref{eq:main_with_p}). If
\begin{eqnarray}
\begin{aligned}
\min_{i}  \frac{\| \beta_i \|_2^p}{\| \bd_i \|_2^p}  \:\min_{i} \| \vec{\bU}_{k_i}^T \dot{\bU}_{k_{i}} \|_{m}^p \ge \frac{\kappa}{\kappa +m-1}   
\frac{\dot{\Delta}_{\max} }{\Delta_{\min}}\:,
  \end{aligned}
  \label{eq:withinnovsuff}
\end{eqnarray}
then  Requirement 1 is satisfied
where $\| \vec{\bU}_{k_i}^T \dot{\bU}_{k_{i}} \|_{m}$ is the minimum singular value of $\vec{\bU}_{k_i}^T \dot{\bU}_{k_{i}}$.
\label{thm:with_inv}
\end{theorem}

The subspace $\Vec{\calS_i}$ was defined as the projection of $\calS_i$ onto $(\oplus_{k \neq i} \calS_k)^{\perp}$ which is equivalent to the projection of $\dot{\calS}_i$ onto $(\oplus_{k \neq i} \calS_k)^{\perp}$. The closer is $\dot{\calS}_i$ to $\Vec{\calS_i}$, the more incoherent is $\dot{\calS_i}$ with the innovative component of the other clusters
since $\vec{\calS_i}$ is orthogonal to $\oplus_{j\neq i} \dot{\calS_j}$. 
 This is the reason we have $\| \vec{\bU}_{k_i}^T \dot{\bU}_{k_{i}} \|_{m}$ on the LHS of (\ref{eq:withinnovsuff}) because $\| \vec{\bU}_{k_i}^T \dot{\bU}_{k_{i}} \|_{m} = \underset{\|u \|_2 =1}{\min} \| \vec{\bU}_{k_i}^T \dot{\bU}_{k_{i}} \bu \|_2$ is a measure of coherence between $\dot{\calS}_i$ and $\vec{\calS}_i$. 
 Therefore, similar to Theorem \ref{thm: 1}, Theorem \ref{thm:with_inv} states that the weaker is the projection of data points onto the innovative components, the more distanced the innovative components should be. The major difference between the condition of  Theorem \ref{thm: 1} and that of Theorem \ref{thm:with_inv} is that in (\ref{eq:withinnovsuff}) $m$ plays a stronger role and (\ref{eq:withinnovsuff}) states that increasing $m$ (provided that it does not increase the coherency between the innovative components) can enhance the chance of MFC/iPursuit to satisfy Requirement 1. The following theorem \textcolor{black}{provides a more explicit sufficient} condition via assuming the random data model used in Theorem \ref{thm:full_random}.
 
\begin{theorem}
Suppose $\bD$ and $\bA$ are generated as in Theorem \ref{thm:full_random} and assume that $M_1 > s + (r-s)m$. 
If 
\begin{eqnarray}
\begin{aligned}
\sigma_{l} \left( \frac{r-s}{r} ,\delta  \right) \sigma _{l}\left( \frac{\vartheta}{M_1} , \delta \right) \ge \frac{\kappa}{\kappa + m - 1} \frac{ \frac{n}{r} + \frac{r-s}{r} {\eta_{\delta}}_{r-s}}{\frac{n}{r} - {\eta_{\delta}}_r }
\end{aligned}
\end{eqnarray}
where $\vartheta = M_1 - \big(s + (r-s)(m-1)\big)$, then Requirement 1 with $p=2$ is satisfied with probability at least $1 - 6\delta - \epsilon$ where $\epsilon$ is the probability that the rank of $\bD$ is less $s + (r-s)m$.
\label{thm:with_inv_random}
\end{theorem}
Note that Theorem \ref{thm:with_inv_random} does not need to explicitly presume that Assumption \ref{asm:with_inv} holds because when $M_1 > s + (r-s)m$, Assumption \ref{asm:with_inv} is satisfied with an overwhelming probability \cite{vershynin2010introduction}.  
The sufficient condition roughly states that when $n/r$ is  large enough, then $\frac{r-s}{s} \: \frac{\vartheta}{M_1}$ 
should be sufficiently larger than $\frac{\kappa}{\kappa + m}$ to guarantee that the requirement is satisfied with high probability.
The value of $\frac{\vartheta}{M_1}$ increases when $M_1$ increases and it converges  to 1 when $r_d/M_1$ decreases.

\begin{figure*}[h!]
\begin{center}
\mbox{
\includegraphics[width=1.38in]{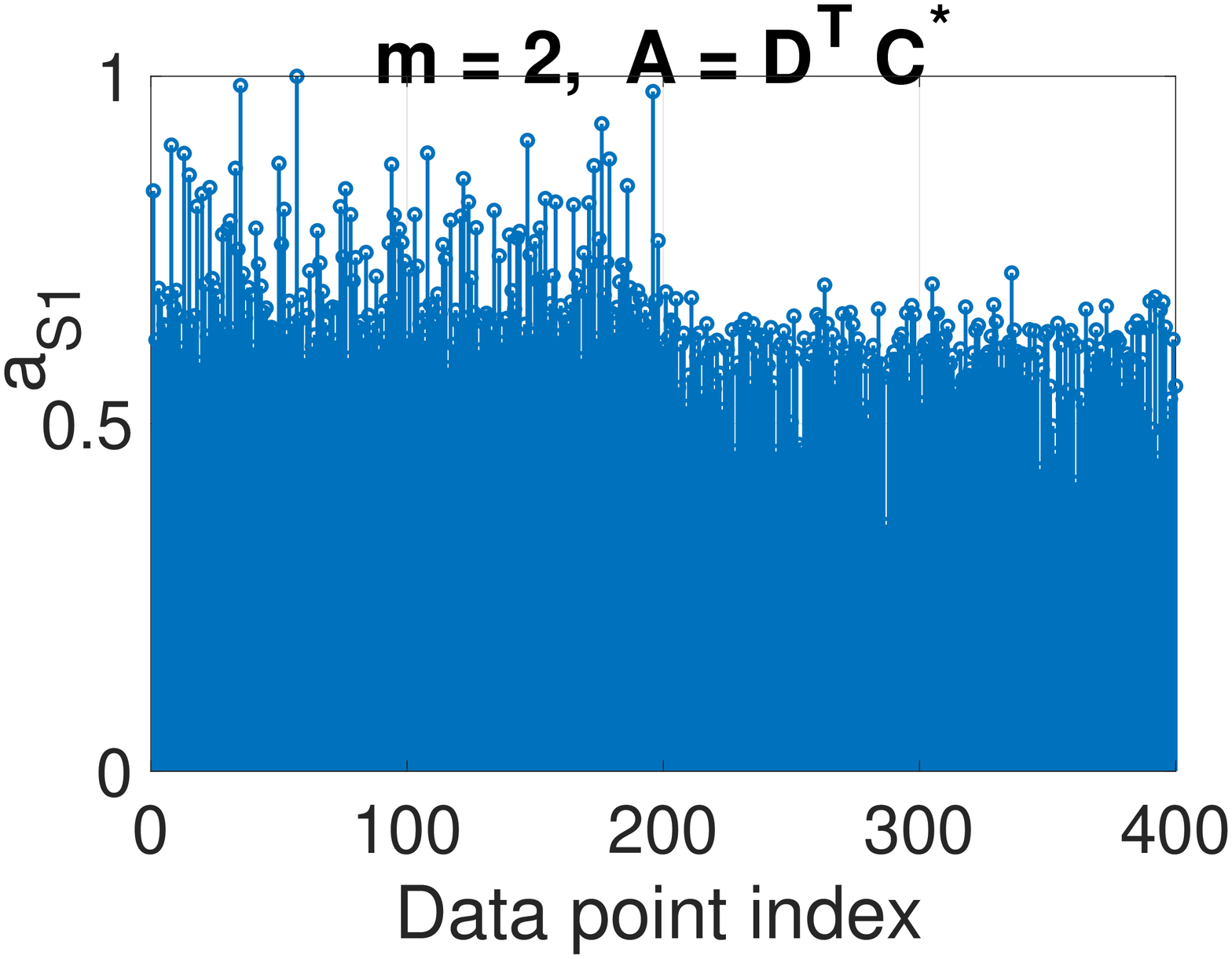}\hspace{-0.1in}
\includegraphics[width=1.38in]{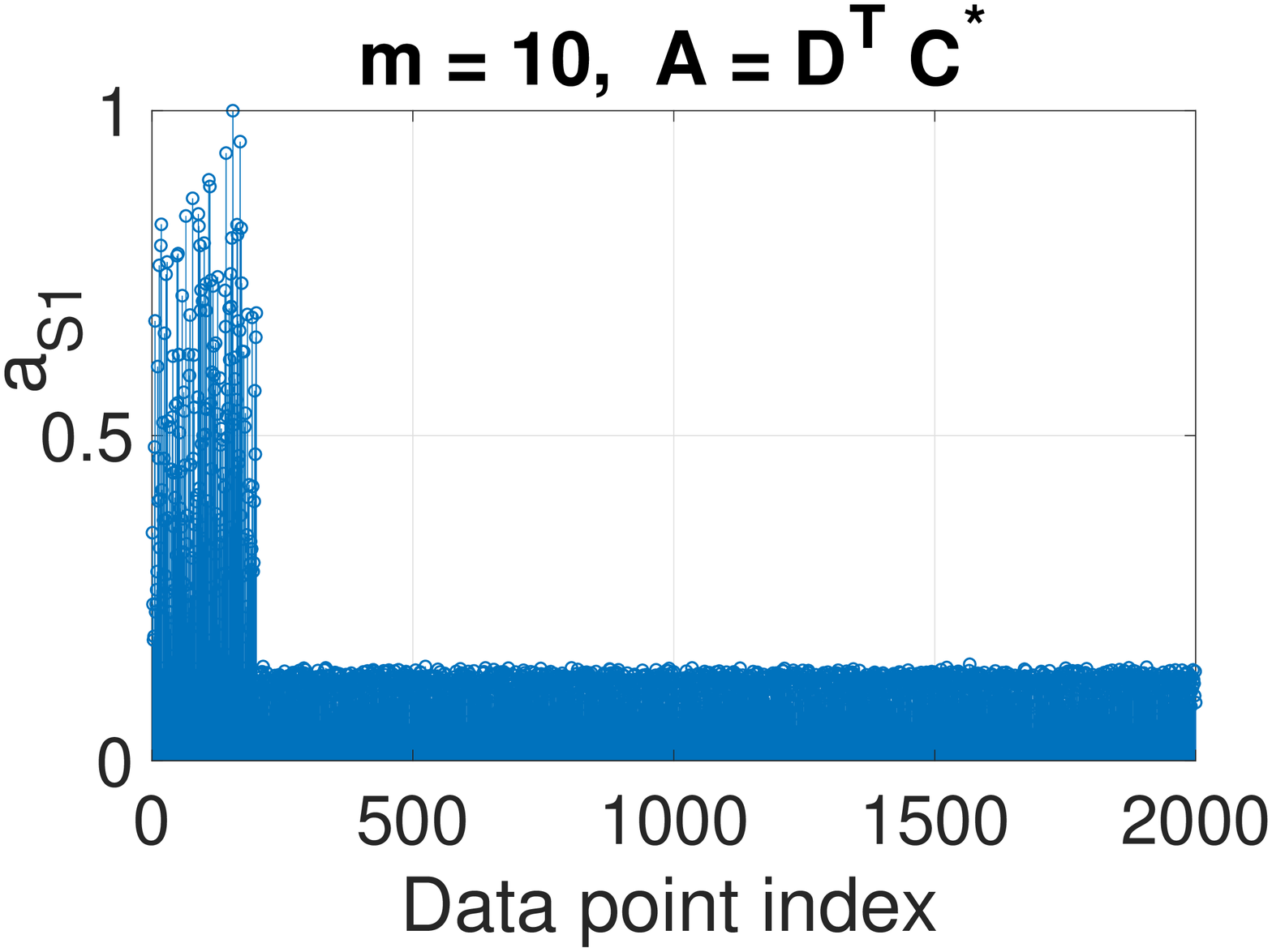}\hspace{-0.1in}
\includegraphics[width=1.38in]{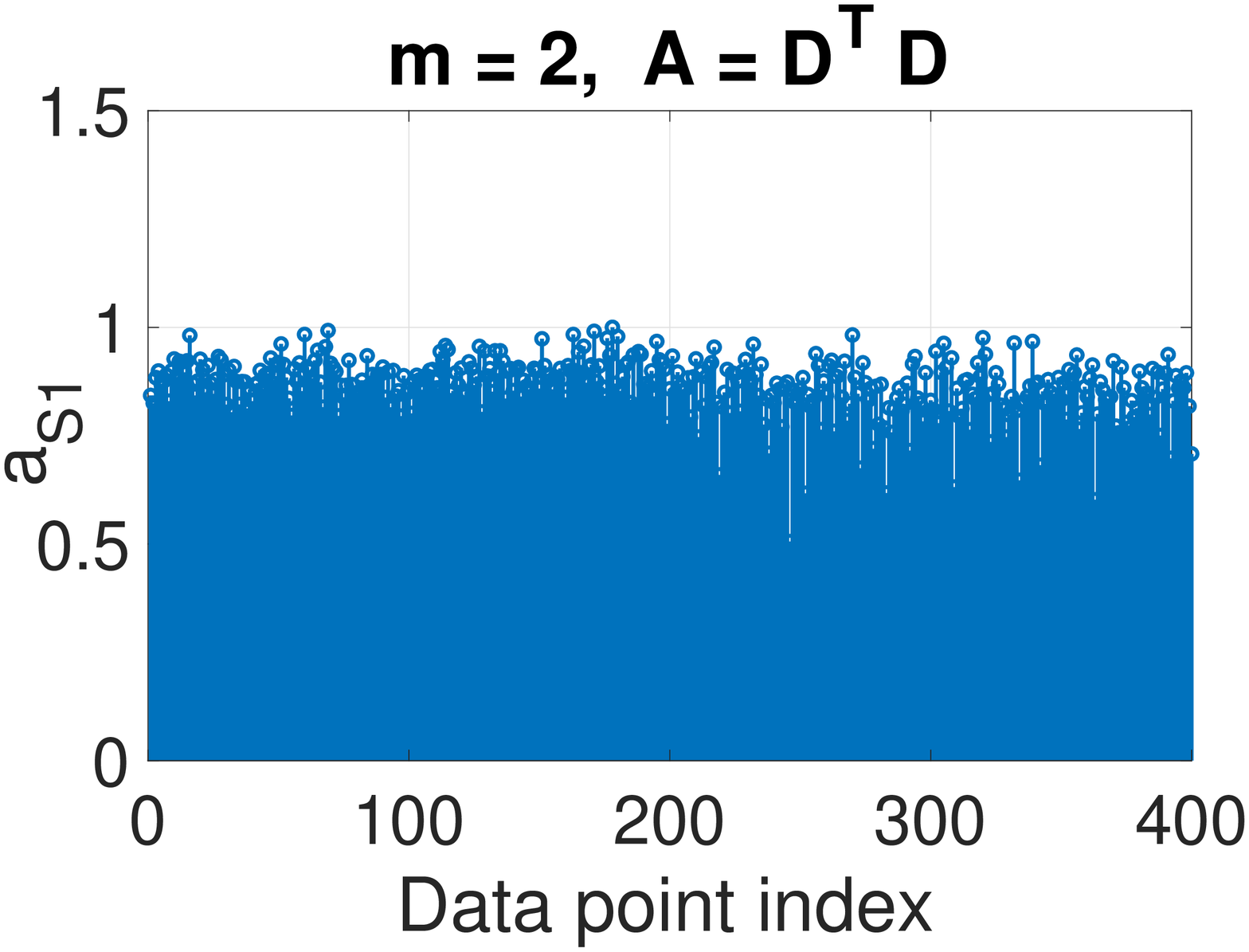}\hspace{-0.1in}
\includegraphics[width=1.38in]{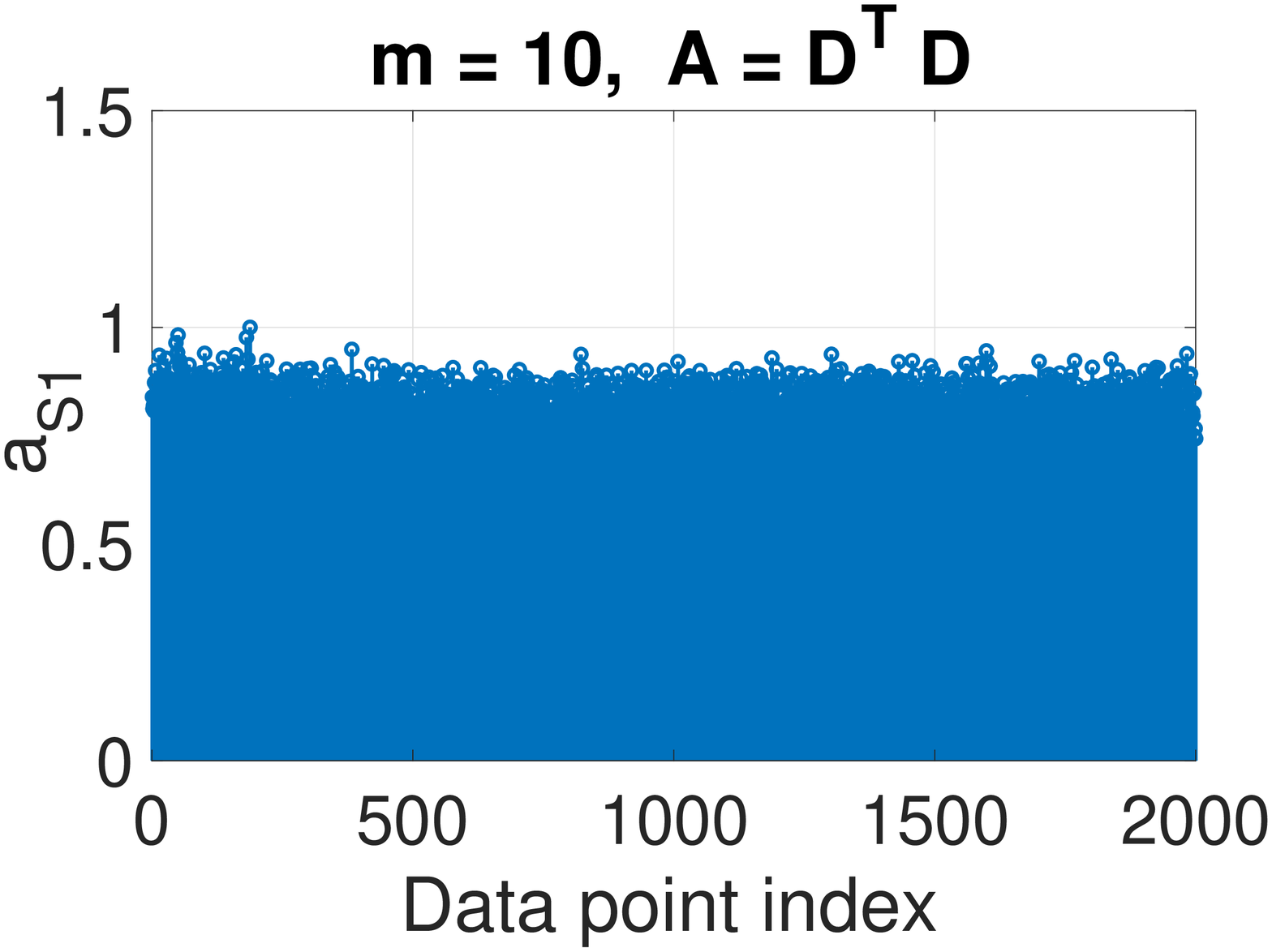}
\hspace{-0.1in}
\includegraphics[width=1.38in]{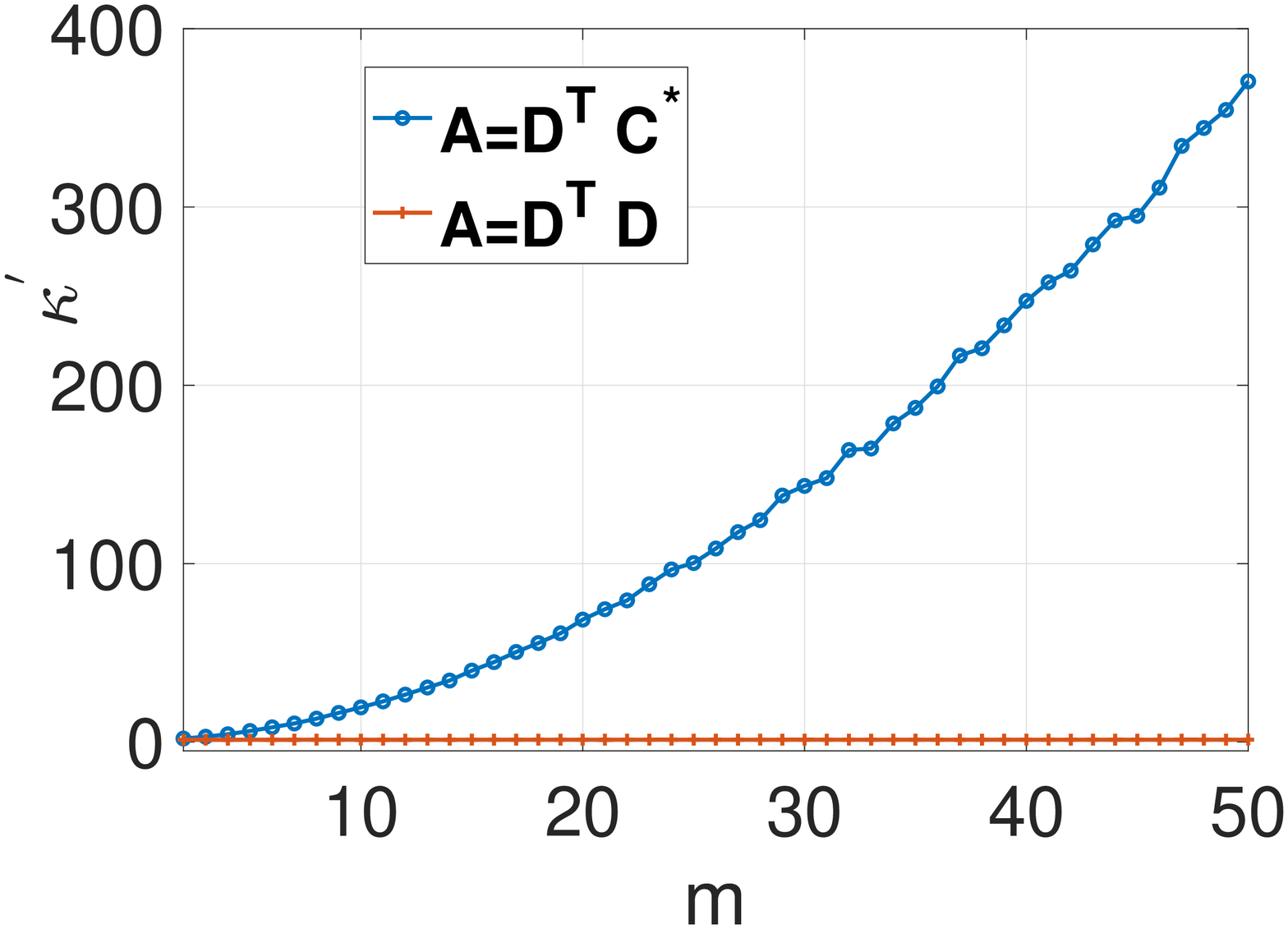}
}
\end{center}
\vspace{-0.15in}
           \caption{The first 4 plots (from LHS) show the elements of $\ba_{\calS_1} =  \frac{1}{n}\sum_{i=1}^{n} \ba_i$ with different number of clusters for MFC and Algorithm 3. The first $n=200$ data points lie in first cluster, $r=10$, $s=8$, and $M_1 = 400$. The last plot demonstrates parameter $\kappa^{'}$ defined in (\ref{eq:define_k_pr}) versus $m$. One can observe that in this experiment increasing $m$ improves the quality of the  adjacency matrix computed by MFC. }
    \label{fig:versus_m_matn}
\end{figure*}

Theorem \ref{thm: 1}, Theorem \ref{thm:with_inv}, and Theorem \ref{thm:with_inv_random} indicate that if $\bD$ follows Data Model 1, then the larger is the number of clusters, the more likely it is for MFC/iPursuit to satisfy Requirement 1 provided that increasing $m$ does not increase the coherency between $\{\dot{\calS}_i \}_{i=1}^m$. This fact might sound counter intuitive, but it is an accurate prediction. For instance, suppose that $\bD$ is generated as in Theorem \ref{thm:with_inv_random}, the first $n$ columns of $\bD$
lie in $\calS_1$, $n=200$, $r=10$, $s=8$, and $M_1=400$. Define 
$
\ba_{\calS_1} = \frac{1}{n}\sum_{i=1}^{n} \ba_i $
where $\ba_i$ is the $i^{th}$ column of $\bA$. Therefore, $\ba_{\calS_1}$ is the average of the first $n$ columns of $\bA$ which are corresponding to data points in $\calS_1$. 
Figure \ref{fig:versus_m_matn} shows $\ba_{\calS_1}$ with different values of $m$ for the adjacency matrices computed by MFC and the TSC algorithm \cite{heckel2013robust} which computes $\bA = |\bD^ T \bD|$. 
Ideally, we should observe that the expected value of the elements of $\ba_{\calS_1}[1:n]$ are  sufficiently larger than the expected value of the elements of $\ba_{\calS_1}[n:M_2]$. 
One can observe that when $\bA = | \bD^T \bD |$, 
the elements of $\ba_{\calS_1}[1:n]$ are  not much distinguishable from the elements of $\ba_{\calS_1}[n:M_2]$ with both $m=2$ and $m=10$.
 In  contrast, when $\bA = | \bD^T \bC^{*} |$ and when $m=10$, $\| \ba_{\calS_1}[1:n] \|_2$ is clearly larger than $\frac{1}{m-1} \| \ba_{\calS_1}[n:M_2]\|_2$. The last plot of Figure \ref{fig:versus_m_matn} shows the effect of $m$ on the quality of the computed adjacency matrix in a more clear way. Define 
parameter $\hat{\kappa}$ as follows
\begin{eqnarray}
\begin{aligned}
{\kappa}^{'} = \frac{(m-1)\: \left\| \ba_{\calS_1}[1:n] \right\|_2^2 }{\left\| \ba_{\calS_1}[n : M_2] \right\|_2^2 } \:.
\end{aligned}
\label{eq:define_k_pr}
\end{eqnarray}
Parameter $\kappa^{'}$ shows how clear the adjacency matrix separates the data points in $\calS_1$ from the other clusters. 
The last plot (first from right), shows ${\kappa}^{'} $ versus $m$ for both MFC and TSC. One can observe that ${\kappa}^{'} $ notably increases as $m$ increases when $\bA = |\bD^T \bC^{*}|$ 
which means that the quality of the estimated adjacency matrix  improves as $m$ increases. In sharp contrast, increasing $m$ does not show a positive/negative impact on the computed adjacency matrix by Algorithm 3.  

It is important to note that the conclusion that the performance of MFC/iPursuit  improves if $m$ increases is not a general rule. When $M_1$ is not sufficiently large, as $m$ increases, the distance between the subspaces (and the distance between their innovative components) decreases and it degrades the performance of the algorithms.  
Moreover,  the reason that in Theorem \ref{thm:full_random} and Theorem \ref{thm:with_inv_random} the coherency between the subspaces decreases as $M_1$ increases is due to the presumed model for the generation of the subspaces and it is not a general rule that $\phi$  decreases as $M_1$ increases.

\begin{algorithm}
\caption{Inner-Product based Subspace Clustering \cite{heckel2013robust} (TSC Algorithm)}
{
\textbf{Input.} The input is data matrix $\bD \in \mathbb{R}^{M_1 \times M_2}$.

\smallbreak
\textbf{1. Data Preprocessing.}  Normalize the $\ell_2$-norm of the columns of $\bD$, i.e., set $\bd_i$ equal to $\bd_i / \| \bd_i \|_2$ for all $1 \le i \le M_2$.

\textbf{2.} Define $\bA = \big| \bD^T \bD|$.

\textbf{3.} Similar to Step 4 in Algorithm 1. 

\textbf{4.} Similar to Step 5 in Algorithm 1. 

\textbf{Output:} The identified clusters.

 }
\end{algorithm}

\subsection{Comparison with  the TSC Algorithm}
\label{sec:compTSC}
In this section, we theoretically compare the TSC algorithm against against MFC/iPursuit. 
 Both MFC/iPursuit and Algorithm 3  use inner-product as the kernel function to measure the similarity between data points. However, in sharp contrast to Algorithm 3,  MFC/iPursuit computes  the inner-product between the directions of innovation  and the data points as opposed to computing the inner-product between the data points. In \cite{rahmani2017innovationJ,rahmani2017subspace} and in this paper, it is shown that this difference makes MFC/iPursuit able to notably outperform TSC in most of scenarios. 
In order to clarify the reason behind this performance difference, we provide similar analysis for Algorithm 3 and we compare the requirements of MFC/iPursuit against those of Algorithm 3. Although the presented theorems only include sufficient conditions (not necessary conditions), their comparison is insightful. 

\begin{theorem}
Suppose $\bD$ follows Data Model 1. If
 \begin{eqnarray}
\begin{aligned}
 & 1 \ge \kappa \max_{i} \left\{ \frac{\| \alpha_i \|_2^p}{\| \bd_i \|_2^p}  \right\}   \frac{\bar{\Delta}_{\max}}{{\Delta}_{\min}}  + \\
 & \qquad \quad\kappa \max_{i} \left\{ \frac{\| \beta_i \|_2^p}{\| \bd_i \|_2^p}  \right\}  \phi^p \frac{\dot{\Delta}_{\max}}{{\Delta}_{\min}} \:,
\end{aligned}
\label{eq:sftc}
\end{eqnarray}
then $\bA = |\bD^T \bD|$ satisfies  Requirement (\ref{asm:sufficeint}). 
\label{thm:tsc_deter}
\end{theorem}
There are two terms on the RHS of the sufficient condition where only the second term is weighted by $\phi$. Even in the best case scenario where the innovative components  are orthogonal to each other, i.e., $\phi = 0$, it may not be possible to satisfy the sufficient condition. For instance, suppose $s/r$ is nearly equal to one and assume that the elements of $\beta_i$ and $\alpha_i$ are sampled independently from $\calN(0,1)$. In this scenario, $\mathbb{E}\left[\frac{\|\alpha_i\|_2^2}{\|\bd_i\|_2^2}\right] = \frac{s}{m} \approx	1$ and it may not be possible to satisfy the sufficient condition even for $\kappa=2$. The main reason is that when $s/m$ is high, the inner-product value between data points in different clusters are high, no matter how well separated the innovative components are.  In sharp contrast to Algorithm 3, MFC/iPursuit utilize the inner-product between the optimal directions and the data points to construct the adjacency matrix and when $s/m$ is high, the optimal directions are strongly incoherent with $\calS$ and this feature  makes the role of the innovative components  notably more significant. In order to make a more explicit comparison, we derive the sufficient condition for Algorithm 3 
while it is assumed that the data is generated as in Theorem \ref{thm:full_random}. The following theorem provides the result. 
\begin{theorem}
Suppose $\bD$ is generated as in Theorem \ref{thm:full_random} and $\bA = | \bD^T \bD |$.  If 
 \begin{eqnarray}
\begin{aligned}
&  \left(\frac{n}{r} - {\eta_{\delta}}_r \right) \ge \kappa \:
\sigma_{u}\left( \frac{s}{r} , \delta \right) 
 \left(\frac{n}{r} + \frac{s}{r} \: {\eta_{\delta}}_s \right) + \\
 & \quad\quad\quad\kappa\: \sigma_{u}\left( \frac{r-s}{r} , \delta \right)  \left(\frac{n}{r} + \frac{r-s}{r} \:{\eta_{\delta}}_{r-s} \right)\left( \frac{c_{\delta} (r-s)^2 }{M_1} \right)\:,
\end{aligned}
 \end{eqnarray}
then 
Requirement \ref{asm:sufficeint} with $p=2$ is satisfied with probability at least $1 - 9\delta$, where $c_{\delta}$ was defined in Theorem \ref{thm:full_random} and ${\eta_{\delta}}_x$ was defined in Theorem \ref{thm:random_point}.
\label{thm:sc_random}
\end{theorem}
The first term on the RHS of the sufficient condition of Theorem \ref{thm:sc_random} is the dominant term when $s$ is large. When there are a sufficiently large number of data points in the clusters ($n/r$ is  large enough), the sufficient condition roughly states that $\kappa \: \frac{s}{r}$ should be \textcolor{black}{sufficiently} smaller than $1$. However,  it is not feasible to satisfy this condition in many scenarios. For instance, if we choose $\kappa = 2$, then the sufficient condition can be satisfied only when $s/r > 0.5$. 

In summary,  comparing the sufficient conditions suggests that in sharp contrast to Algorithm 3 which fails when the span of clusters are close,  MFC/iPursuit can effectively leverage the innovative components of the clusters and if these innovative components are sufficiently separable ($\phi$ is sufficiently small), MFC/iPursuit might successfully distinguish the clusters. 

\begin{figure*}[h!]
\begin{center}
\mbox{
\includegraphics[width=1.6in]{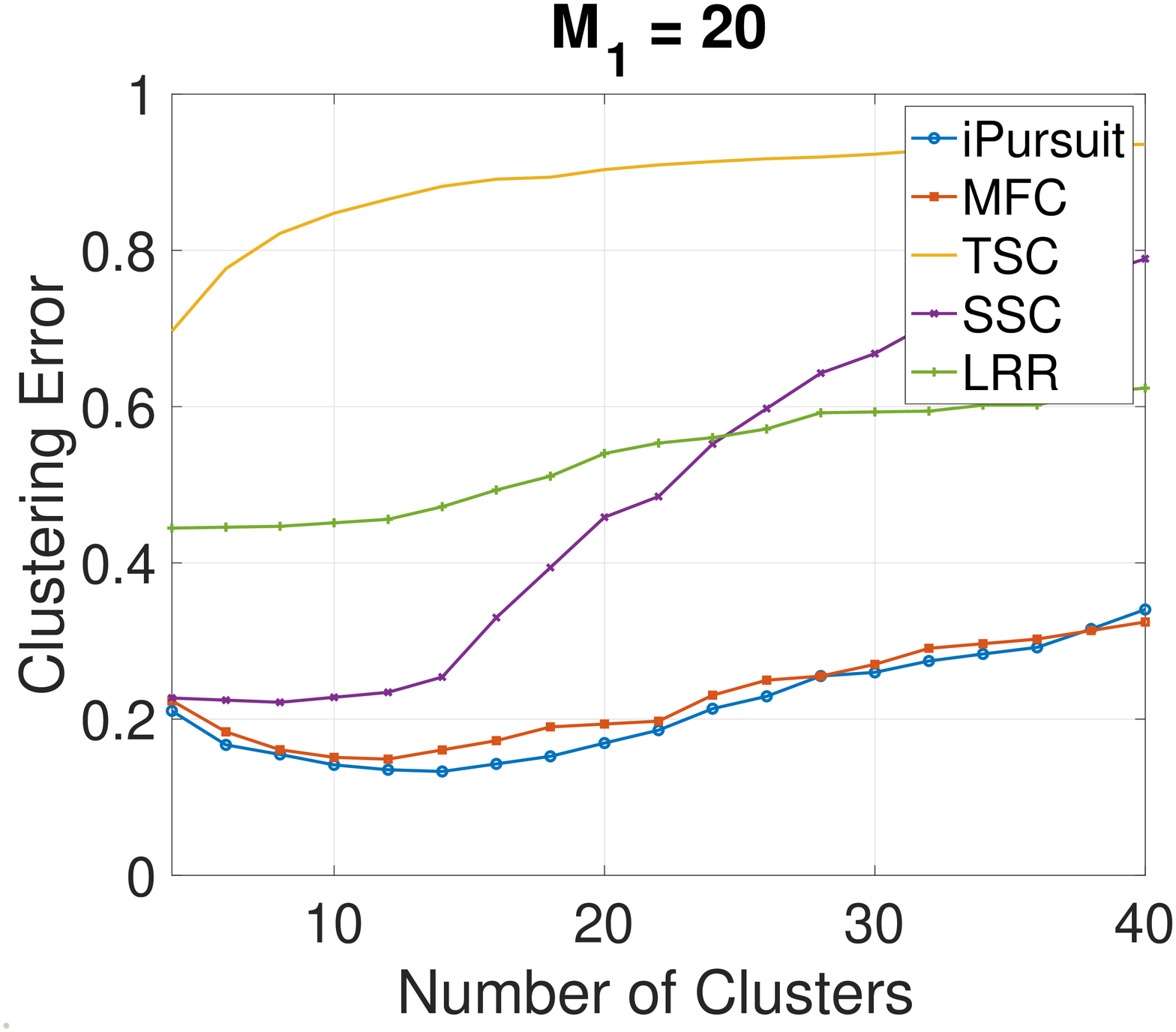}\hspace{-0.1in}
\includegraphics[width=1.6in]{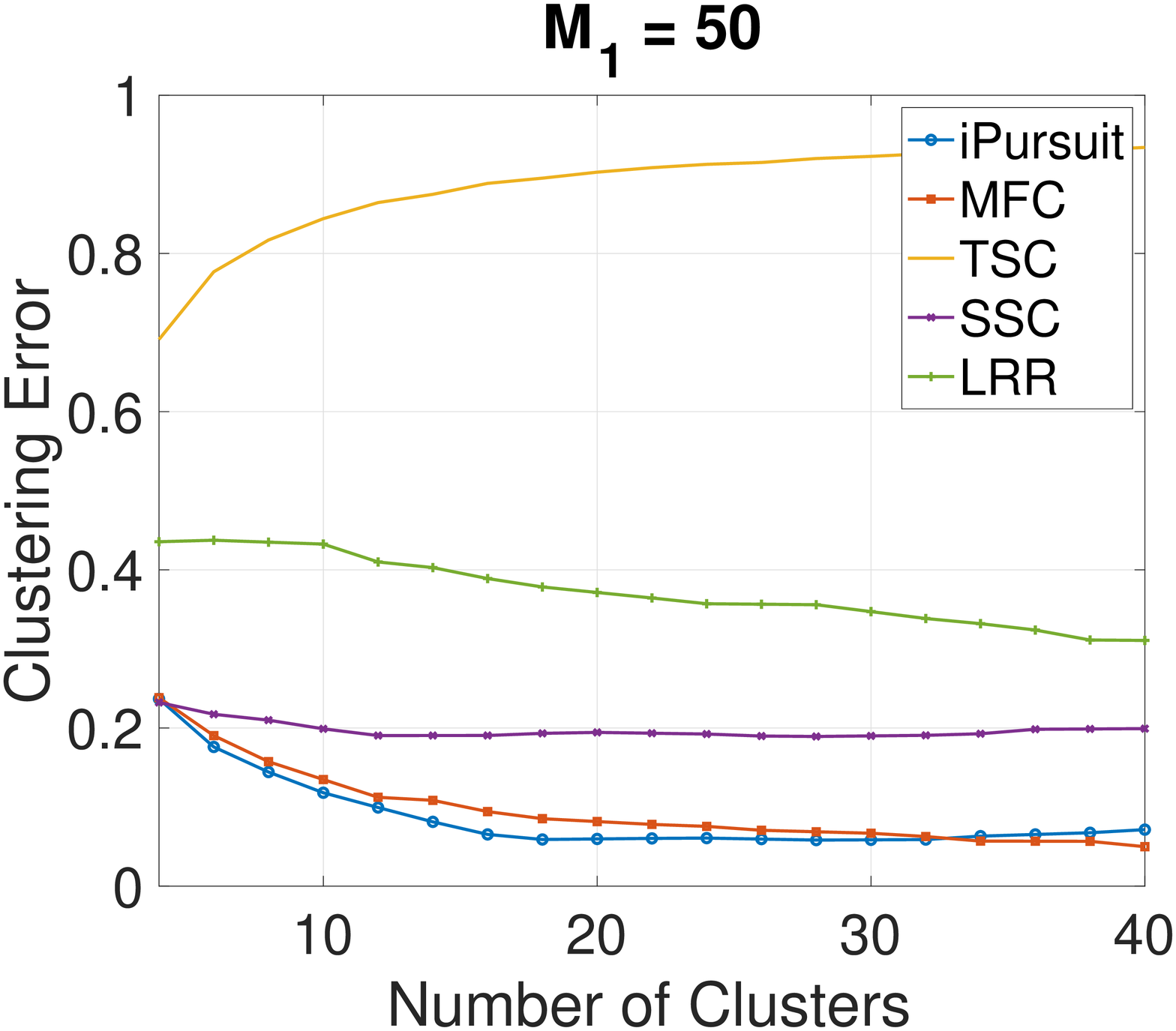}\hspace{-0.1in}
\includegraphics[width=1.6in]{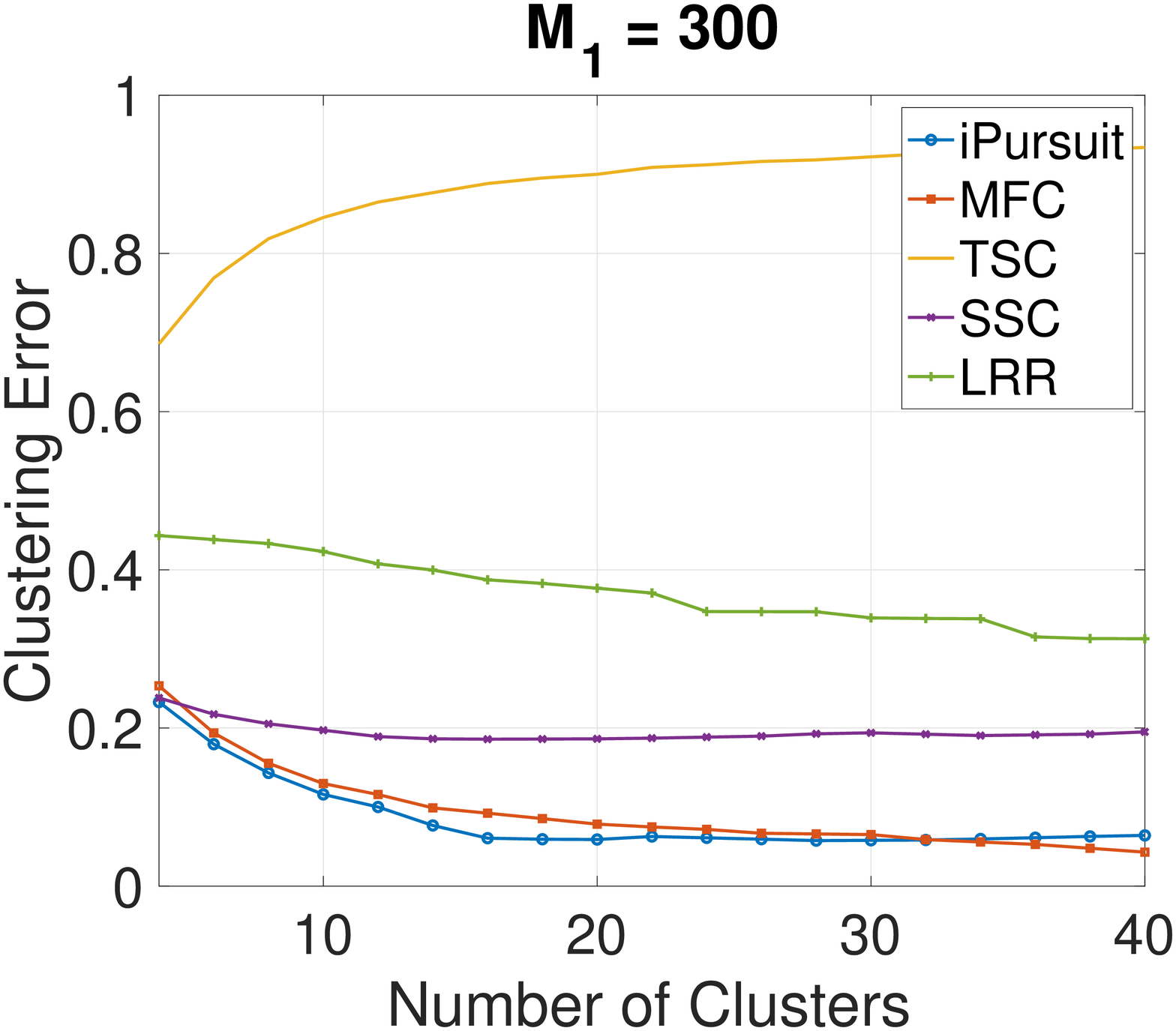}
\hspace{0.1in}
\includegraphics[width=1.6in]{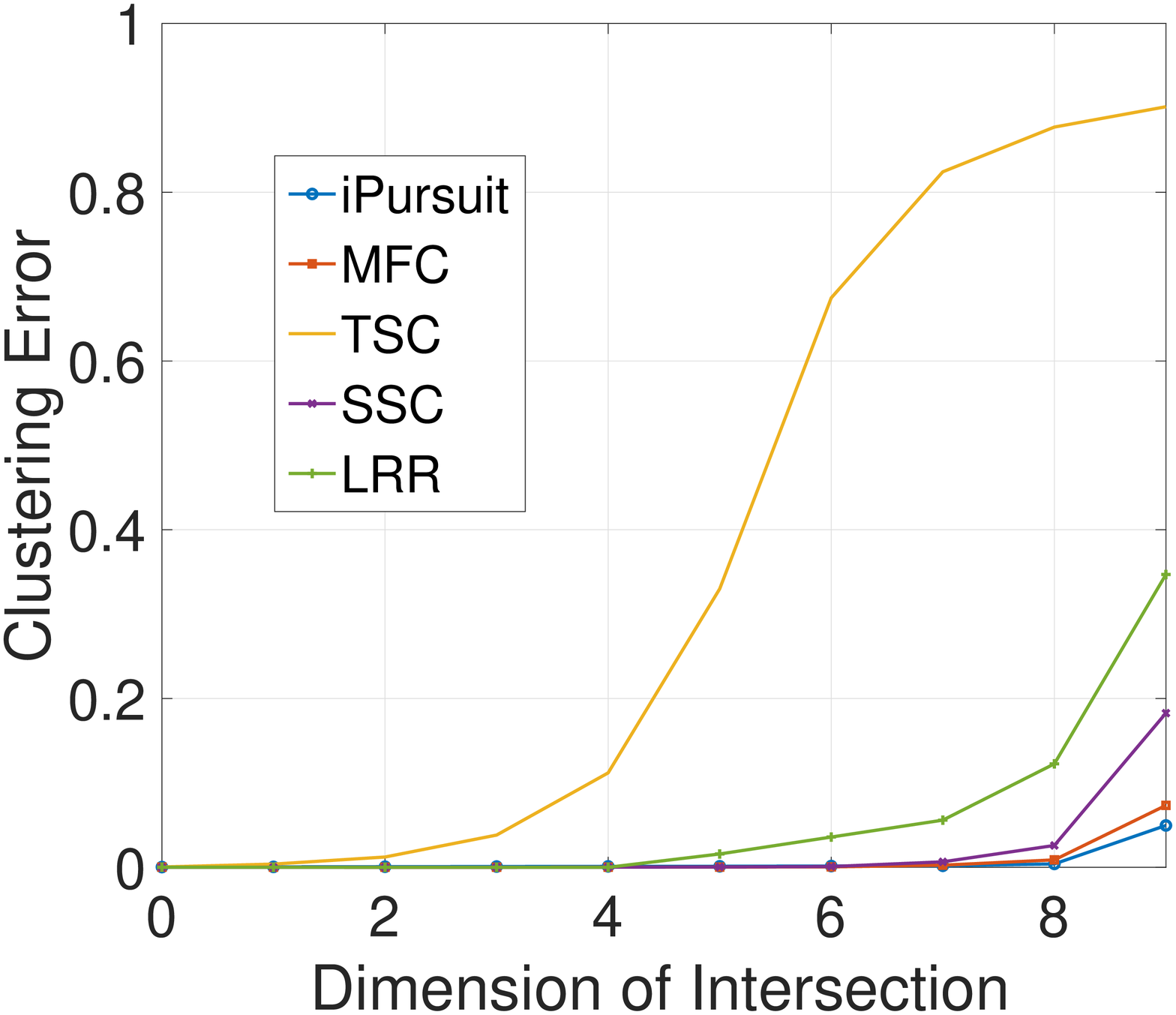}
}
\end{center}
\vspace{-0.15in}
           \caption{First three plots from left: Clustering error versus the number of clusters for different values of $M_1$ where  $r=10$, $s=9$, and $n=100$. First plot from right: This plot demonstrates clustering error versus $s$. In this experiment, $M_1=40$, $r=10$, and $n=100$.}
    \label{fig:versus_m}
\end{figure*}

\section{Numerical Experiments}
This paper does not present a new clustering algorithm and the main focus was to provide a deep understating and analysis of the MFC/iPursuit algorithms. We refer the reader to \cite{kanatani2001motion,costeira1998multibody,boult1991factorization,vidal2011subspace,rahmani2017innovation,rahmani2017innovationJ} for numerical studies of the MFC/iPursuit algorithms. The focus of the presented experiments are to demonstrate some of the features of the algorithms which was predicted by the presented theoretical studies. 
 For iPursuit, MFC, and TSC, the graph preprocessing step (Step 4 in Algorithm 1) was done as follows. For each column of $\bA$, 8 largest elements were kept and the rest of elements were set to zero.   The clustering error  is defined as $\frac{N_e}{M_2}$ where $N_e$ is the total number of misclassified data points. In the appendix, we have included a simple numerical experiment showing that exact clustering can be achieved if Requirement 1 holds even for small values of $\kappa$. 

\subsection{The Dimension of Intersection Between the Subspaces}
In the presented deterministic results (Theorem \ref{thm: 1} and Theorem \ref{thm:with_inv}), we observed that $\frac{\| \beta_i \|_2}{\| \bd_i \|_2}$ is an important factor in the performance of MFC/iPursuit and in the probabilistic results, this factor appeared as $\frac{r-s}{r}$.
The purpose of this experiment is twofold. Firstly, we show that the accuracy of MFC/iPursuit degrades as $s$ increases (since $\frac{r-s}{r}$ decreases). Secondly, it is shown that MFC/iPursuit are notably robust against intersection between the span of  clusters comparing to most of other methods. The first plot (from right) in Figure \ref{fig:versus_m} shows clustering error versus $s$ where in this experiment $M_1=40$, $r=10$, and $n=100$ (the number of evaluation
runs was 50). One can observe that the accuracy of MFC/iPursuit degrades as $s$ increases. However, both of them notably outperform the other methods when $s$ is high. The main reason is that as the presented theoretical studies  indicated, the performance of MFC/iPursuit mainly depends on the coherency between the innovative components $\{ \dot{\calS}_i \}_{i=1}^m$ while most of other algorithms such as TSC require the span of clusters $\{ {\calS}_i \}_{i=1}^m$ to be sufficiently incoherent.  

\subsection{Number of Clusters}
\label{sec:exp_m}
In the theoretical results (Theorem \ref{thm: 1} and Theorem \ref{thm:with_inv}), it was shown that the quality of the adjacency matrix computed by MFC/iPursuit might improve when $m$ increases. Specifically, the theoretical results suggested that when data follows Data Model 1 and as long as increasing $m$ does not increase the coherency between $\{\dot{\calS}_{i}\}_{i=1}^{m}$, MFC/iPursuit can yield  a better adjacency matrix (an adjacency matrix with higher $ \min_i \frac{(m-1)\| {\ba_i}_{\calI_i} \|_p^p  }{ \| {\ba_i}_{\calI_i^{\perp}} \|_p^p }$)  if $m$ increases.

The first three plots (from left) in Figure \ref{fig:versus_m} shows clustering error versus $m$ for different values of $M_1$ where in this experiment $r=10$, $s=9$, and $n=100$ (the number of evaluation
runs was 50). One can observe that when $M_1 = 300$ and when $M_1=50$, the accuracy of MFC/iPursuit improves when $m$ increases while when $M_1 = 20$, the accuracy degrades. The reason for this observation is that as the theoretical results indicated,  both the number of clusters and the coherency between the innovative components contribute to the performance of the algorithms. When $M_1$ is not sufficiently large, increasing $m$ increases the coherency between the innovative components and it degrades the performances of the algorithms.

\subsection*{Conclusion}
It was shown that  iPursuit  is equivalent to a closed form matrix factorization based clustering algorithm if the direction search optimization problem is altered into a quadratic optimization problem. A novel analysis applicable to both algorithms were proposed which showed that in contrast to some of the other subspace clustering algorithms whose performance depend on the distance between the span of clusters, the performance of MFC/iPursuit mainly depends on the distance between the innovative components of the clusters. 

\newpage
 \bibliography{bibfile}

\begin{thebibliography}{10}

\bibitem{boult1991factorization}
Terrance~E Boult and L~Gottesfeld Brown.
\newblock Factorization-based segmentation of motions.
\newblock In {\em Proceedings of the IEEE workshop on visual motion}, pages
  179--180. IEEE Computer Society, 1991.

\bibitem{bradley2000k}
Paul~S Bradley and Olvi~L Mangasarian.
\newblock k-plane clustering.
\newblock {\em Journal of Global Optimization}, 16(1):23--32, 2000.

\bibitem{candes2009exact}
Emmanuel~J Cand{\`e}s and Benjamin Recht.
\newblock Exact matrix completion via convex optimization.
\newblock {\em Foundations of Computational mathematics}, 9(6):717, 2009.

\bibitem{chen2009spectral}
Guangliang Chen and Gilad Lerman.
\newblock Spectral curvature clustering ({SCC}).
\newblock {\em International Journal of Computer Vision}, 81(3):317--330, 2009.

\bibitem{costeira1998multibody}
Jo{\~a}o~Paulo Costeira and Takeo Kanade.
\newblock A multibody factorization method for independently moving objects.
\newblock {\em International Journal of Computer Vision}, 29(3):159--179, 1998.

\bibitem{dyer2013greedy}
Eva~L Dyer, Aswin~C Sankaranarayanan, and Richard~G Baraniuk.
\newblock Greedy feature selection for subspace clustering.
\newblock {\em The Journal of Machine Learning Research}, 14(1):2487--2517,
  2013.

\bibitem{elhamifar2013sparse}
Ehsan Elhamifar and Rene Vidal.
\newblock Sparse subspace clustering: Algorithm, theory, and applications.
\newblock {\em IEEE Transactions on Pattern Analysis and Machine Intelligence},
  35(11):2765--2781, 2013.

\bibitem{feng2014robust}
Jiashi Feng, Zhouchen Lin, Huan Xu, and Shuicheng Yan.
\newblock Robust subspace segmentation with block-diagonal prior.
\newblock In {\em Proceedings of the IEEE conference on computer vision and
  pattern recognition}, pages 3818--3825, 2014.

\bibitem{rnc1}
Martin~A Fischler and Robert~C Bolles.
\newblock Random sample consensus: a paradigm for model fitting with
  applications to image analysis and automated cartography.
\newblock {\em Communications of the ACM}, 24(6):381--395, 1981.

\bibitem{gao2015multi}
Hongchang Gao, Feiping Nie, Xuelong Li, and Heng Huang.
\newblock Multi-view subspace clustering.
\newblock In {\em Proceedings of the IEEE International Conference on Computer
  Vision}, pages 4238--4246, 2015.

\bibitem{heckel2013robust}
Reinhard Heckel and Helmut B{\"o}lcskei.
\newblock Robust subspace clustering via thresholding.
\newblock {\em arXiv preprint arXiv:1307.4891}, 2013.

\bibitem{ji2017deep}
Pan Ji, Tong Zhang, Hongdong Li, Mathieu Salzmann, and Ian Reid.
\newblock Deep subspace clustering networks.
\newblock {\em Advances in neural information processing systems}, 30:24--33,
  2017.

\bibitem{jiang2018nonconvex}
Hao Jiang, Daniel~P Robinson, Rene Vidal, and Chong You.
\newblock A nonconvex formulation for low rank subspace clustering: algorithms
  and convergence analysis.
\newblock {\em Computational Optimization and Applications}, 70(2):395--418,
  2018.

\bibitem{junge2013noncommutative}
Marius Junge, Qiang Zeng, et~al.
\newblock Noncommutative bennett and rosenthal inequalities.
\newblock {\em The Annals of Probability}, 41(6):4287--4316, 2013.

\bibitem{kanatani2001motion}
Ken-ichi Kanatani.
\newblock Motion segmentation by subspace separation and model selection.
\newblock In {\em Proceedings Eighth IEEE International Conference on computer
  Vision. ICCV 2001}, volume~2, pages 586--591. IEEE, 2001.

\bibitem{klys2018learning}
Jack Klys, Jake Snell, and Richard Zemel.
\newblock Learning latent subspaces in variational autoencoders.
\newblock In {\em Advances in Neural Information Processing Systems}, pages
  6444--6454, 2018.

\bibitem{laurent2000adaptive}
Beatrice Laurent and Pascal Massart.
\newblock Adaptive estimation of a quadratic functional by model selection.
\newblock {\em Annals of Statistics}, pages 1302--1338, 2000.

\bibitem{ledoux2005concentration}
Michel Ledoux.
\newblock {\em The concentration of measure phenomenon}.
\newblock Number~89. American Mathematical Soc., 2005.

\bibitem{lerman2015robust}
Gilad Lerman, Michael~B McCoy, Joel~A Tropp, and Teng Zhang.
\newblock Robust computation of linear models by convex relaxation.
\newblock {\em Foundations of Computational Mathematics}, 15(2):363--410, 2015.

\bibitem{li2021provable}
Weiwei Li, Mostafa Rahmani, and Ping Li.
\newblock Provable data clustering via innovation search.
\newblock {\em arXiv preprint arXiv:2108.06888}, 2021.

\bibitem{ling2020certifying}
Shuyang Ling and Thomas Strohmer.
\newblock Certifying global optimality of graph cuts via semidefinite
  relaxation: A performance guarantee for spectral clustering.
\newblock {\em Foundations of Computational Mathematics}, 20(3):367--421, 2020.

\bibitem{lipor2021subspace}
John Lipor, David Hong, Yan~Shuo Tan, and Laura Balzano.
\newblock Subspace clustering using ensembles of k-subspaces.
\newblock {\em Information and Inference: A Journal of the IMA}, 10(1):73--107,
  2021.

\bibitem{liu2013robust}
Guangcan Liu, Zhouchen Lin, Shuicheng Yan, Ju~Sun, Yong Yu, and Yi~Ma.
\newblock Robust recovery of subspace structures by low-rank representation.
\newblock {\em IEEE Transactions on Pattern Analysis and Machine Intelligence},
  35(1):171--184, 2013.

\bibitem{lu2013correlation}
Canyi Lu, Jiashi Feng, Zhouchen Lin, and Shuicheng Yan.
\newblock Correlation adaptive subspace segmentation by trace lasso.
\newblock In {\em Proceedings of the IEEE international conference on computer
  vision}, pages 1345--1352, 2013.

\bibitem{menon2020subspace}
Vishnu Menon, Gokularam Muthukrishnan, and Sheetal Kalyani.
\newblock Subspace clustering without knowing the number of clusters: A
  parameter free approach.
\newblock {\em IEEE Transactions on Signal Processing}, 68:5047--5062, 2020.

\bibitem{park2014greedy}
Dohyung Park, Constantine Caramanis, and Sujay Sanghavi.
\newblock Greedy subspace clustering.
\newblock In {\em Advances in Neural Inf. Processing Systems}, pages
  2753--2761, 2014.

\bibitem{patel2013latent}
Vishal~M Patel, Hien Van~Nguyen, and Rene Vidal.
\newblock Latent space sparse subspace clustering.
\newblock In {\em Proceedings of the IEEE international conference on computer
  vision}, pages 225--232, 2013.

\bibitem{peng2016deep}
Xi~Peng, Shijie Xiao, Jiashi Feng, Wei-Yun Yau, and Zhang Yi.
\newblock Deep subspace clustering with sparsity prior.
\newblock In {\em IJCAI}, pages 1925--1931, 2016.

\bibitem{rahmani2017innovation}
Mostafa Rahmani and George Atia.
\newblock Innovation pursuit: A new approach to the subspace clustering
  problem.
\newblock In {\em Proceedings of the International Conference on Machine
  Learning (ICML)}, pages 2874--2882, 2017.

\bibitem{rahmani2017innovationJ}
Mostafa Rahmani and George~K Atia.
\newblock Innovation pursuit: A new approach to subspace clustering.
\newblock {\em IEEE Transactions on Signal Processing}, 65(23):6276--6291,
  2017.

\bibitem{rahmani2017subspace}
Mostafa Rahmani and George~K Atia.
\newblock Subspace clustering via optimal direction search.
\newblock {\em IEEE Signal Processing Letters}, 24(12):1793--1797, 2017.

\bibitem{soltanolkotabi2012geometric}
Mahdi Soltanolkotabi, Emmanuel~J Candes, et~al.
\newblock A geometric analysis of subspace clustering with outliers.
\newblock {\em The Annals of Statistics}, 40(4):2195--2238, 2012.

\bibitem{stat2}
Yasuyuki Sugaya and Kenichi Kanatani.
\newblock Geometric structure of degeneracy for multi-body motion segmentation.
\newblock In {\em Statistical Methods in Video Processing}, pages 13--25.
  Springer, 2004.

\bibitem{stat1}
Michael~E Tipping and Chris~M Bishop.
\newblock Mixtures of probabilistic principal component analyzers.
\newblock {\em Neural computation}, 11(2):443--482, 1999.

\bibitem{DBLduallll}
Manolis~C. Tsakiris and Rene Vidal.
\newblock Hyperplane clustering via dual principal component pursuit.
\newblock In {\em International Conference on Machine Learning, {ICML}},
  volume~70, pages 3472--3481, 2017.

\bibitem{vershynin2010introduction}
Roman Vershynin.
\newblock Introduction to the non-asymptotic analysis of random matrices.
\newblock {\em arXiv preprint arXiv:1011.3027}, 2010.

\bibitem{vidal2011subspace}
Rene Vidal.
\newblock Subspace clustering.
\newblock {\em IEEE Signal Processing Magazine}, 2(28):52--68, 2011.

\bibitem{vidal2005generalized}
Rene Vidal, Yi~Ma, and Shankar Sastry.
\newblock Generalized principal component analysis ({GPCA}).
\newblock {\em IEEE Transactions on Pattern Analysis and Machine Intelligence},
  27(12):1945--1959, 2005.

\bibitem{von2007tutorial}
Ulrike Von~Luxburg.
\newblock A tutorial on spectral clustering.
\newblock {\em Statistics and computing}, 17(4):395--416, 2007.

\bibitem{wang2016noisy}
Yu-Xiang Wang and Huan Xu.
\newblock Noisy sparse subspace clustering.
\newblock {\em The Journal of Machine Learning Research}, 17(1):320--360, 2016.

\bibitem{wang2013provable}
Yu-Xiang Wang, Huan Xu, and Chenlei Leng.
\newblock Provable subspace clustering: When lrr meets ssc.
\newblock {\em Advances in Neural Information Processing Systems}, 26:64--72,
  2013.

\bibitem{yang2006robust}
Allen~Y Yang, Shankar~R Rao, and Yi~Ma.
\newblock Robust statistical estimation and segmentation of multiple subspaces.
\newblock In {\em Computer Vision and Pattern Recognition Workshop (CVPRW)},
  pages 99--99, 2006.

\bibitem{you2016scalable}
Chong You, Daniel Robinson, and Rene Vidal.
\newblock Scalable sparse subspace clustering by orthogonal matching pursuit.
\newblock In {\em Proceedings of the IEEE conference on computer vision and
  pattern recognition}, pages 3918--3927, 2016.

\bibitem{zhang2018cappronet}
Liheng Zhang, Marzieh Edraki, and Guo-Jun Qi.
\newblock Cappronet: Deep feature learning via orthogonal projections onto
  capsule subspaces.
\newblock In {\em Advances in Neural Information Processing Systems}, pages
  5814--5823, 2018.

\bibitem{zhang2014novel}
Teng Zhang and Gilad Lerman.
\newblock A novel m-estimator for robust {PCA}.
\newblock {\em J. of Machine Learning Research}, 15(1):749--808, 2014.

\end{thebibliography}
 \bibliographystyle{plain}



\newpage
\appendix

\section{Appendix}

\subsection{Requirement 1}
\begin{figure}[h!]
\begin{center}
\mbox{
\includegraphics[width=0.35\textwidth]{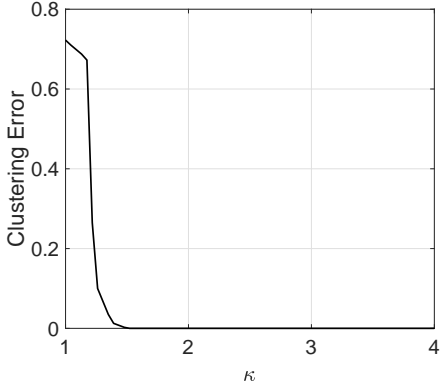}
}
\end{center}
\vspace{-0.15in}
           \caption{Clustering error versus parameter $\kappa$. }
\label{fig:app}
\end{figure}

We discussed the fact that Requirement 1 indicates how clear the estimated adjacency matrix represents the clustering structure of the data and it is similar to the sufficient condition established in \cite{ling2020certifying} which guarantees that the spectral clustering algorithm can yield exact clustering.  In this experiment, we assume that $m=4$ and $n=100$ which means $M_2 = 400$. In order to construct $\bA$, we sample each element of $\bA$ from half-normal distribution and we normalize the elements such that $
 \frac{\kappa}{m-1} \| {\ba_i}_{\calI_i^{\perp}} \|_1 = \| {\ba_i}_{\calI_i} \|_1  \:,  
$ for all $1 \le i \le M_2$. Figure \ref{fig:app} shows clustering error of the spectral clustering algorithm versus $\kappa$. One can observe that even a small value of $\kappa$ can guarantee exact clustering.  Although the minimum value of $\kappa$ for which we can guarantee exact clustering depends on the distribution of the elements of $\bA$, but it shows that (as the results in \cite{ling2020certifying} suggests), exact clustering can be achieved if the false connections are sufficiently weaker than the true connections.

\newpage
\section{Proof of the Presented Theoretical Results}
In this section, the proofs for the presented results are presented. 
\\
\\
\textbf{Proof of Lemma \ref{lm:sade}}\\
The optimal point of (\ref{eq:ell2}) is equivalent to the optimal point of \begin{eqnarray}
\underset{ \bc}{\min} \: \:  \bc^T \bD\bD^T \bc \quad \text{subject to} \quad \bc^T \bd_i = 1 \:,
\label{eq:bflag}
\end{eqnarray}
whose Lagrangian function  is as follows
\begin{eqnarray}
\bc^T \bD\bD^T \bc + \gamma( \bc^T \bd_i - 1) \:,
\end{eqnarray}
where $\gamma$ is the Lagrangian multiplier. 
Therefore, an optimal solution of (\ref{eq:ell2}) which lies in the column space of $\bD$ is equal to 
\begin{eqnarray}
\begin{aligned}
    \bc_i^{*} = \frac{ \bU \Sigma^{-2} \bU^T \: \bd_i}{\bd_i^T \: \bU \Sigma^{-2} \bU^T \: \bd_i} \:.
\end{aligned}
\label{eq:cistar}
\end{eqnarray}
In addition, $\bd_i = \bU \Sigma \bv^i$ where $\bv^i$ is the $i^{th}$ row of $\bV \in \mathbb{R}^{M_2 \times r_d}$. Accordingly,
\begin{eqnarray*}
\begin{aligned}
    \bc_i^{*} = \frac{ \bU \Sigma^{-1} \bv^i }{\|\bv^i \|_2^2}
\end{aligned}
\end{eqnarray*}
and $\bd_j^T \bc_i^{*} = \frac{{\bv^j}^T \bv^i}{\|\bv^i \|_2^2}$.
\\
\\
\textbf{Proof of Theorem \ref{thm: 1}}\\
In the MFC/iPursuit method, the $i^{th}$ column of the estimated adjacency is equal to $\bD^T \bc_i^{*}$.
In order to guarantee that Requirement \ref{asm:sufficeint} holds, it is sufficient to show that 
\begin{eqnarray}
\| \bD_{k_i}^T \bc_i^{*} \|_p^p > \frac{\kappa}{\kappa + (m-1)} \| \bD^T \bc_i^{*} \|_p^p
\label{eq:th1main}
\end{eqnarray} 
holds for all the columns. In order to guarantee that this inequality holds, we establish a lower-bound for the LHS  and an upper-bound for the RHS and we derive the final sufficient condition such that the lower-bound is larger than the upper-bound.

First we derive a lower bound for $ \| \bD_{k_i}^T \bc_i^{*} \|_p^p$. 
According to the linear constraint,
$${\bc_i^{*}}^T {\bd_i} = 1\:,$$
and since $\bd_i \in \calS_{k_i}$, we conclude that
\begin{eqnarray}
\| \bU_{k_i}^T \bc_i^{*} \|_2 \ge \frac{1}{\|\bd_i \|_2} \:.
\label{eq:firstinq}
\end{eqnarray}
All the columns of $\bD_{k_i}$ lie in $\calS_{k_i}$. Thus,
\begin{eqnarray}
  \begin{aligned}
 \| \bD_{k_i}^T \bc_i^{*} \|_p & =  \| \bD_{k_i}^T \bU_{k_i} \bU_{k_i}^T \bc_i^{*} \|_p \:,
\label{eq:lower1}
   \end{aligned}
\end{eqnarray}
and according to (\ref{eq:firstinq}), 
\begin{eqnarray}
 \| \bD_{k_i}^T \bc_i^{*} \|_p^p \ge  \frac{1}{\|\bd_i \|_2^p} \underset{\|\bu\| = 1 \atop \bu \in \calS_{k_i}}{\inf} \:\| \bu^T \bD_{k_i} \|_p^p \:.
 \label{eq:neww}
\end{eqnarray}
Using (\ref{eq:neww}) we  can derive the following lower-bound 
\begin{eqnarray}
   \begin{aligned}
  \| \bD_{k_i}^T \bc_i^{*} \|_p^p & \ge   \frac{1}{\|\bd_i \|_2^p} \min_{j} \left\{ \underset{\|\bu\| = 1 \atop \bu \in \calS_{j}}{\inf} \:\| \bu^T \bD_{j} \|_p^p \right\}_{j=1}^{m} \\
  & = \frac{1}{\|\bd_i \|_2^p} \: \Delta_{\min} \:.
   \end{aligned}
   \label{eq:lhs}
\end{eqnarray}

Next, we establish an upper-bound for the RHS of (\ref{eq:th1main}). Corresponding to each data point $\bd_i$, we define vector $\bd_i^{\perp}$ as 
\begin{eqnarray}
\bd_i^{\perp} = \frac{\dU_{k_i} \beta_i}{\| \beta_i \|_2^2 } \:,
\end{eqnarray}
where $\beta_i \in \mathbb{R}^{r-s}$ was defined in (\ref{eq:alpha_beta}).
Note that 
$$
\bd_i^T \bd_i^{\perp} = \frac{(\bS \alpha_i + \dU_{k_i} \beta_i)^T \dU_{k_i} \beta_i}{\| \beta_i \|_2^2 } = 1 \:.
$$
Since $\bd_i^{\perp} $ is in the feasible set of \begin{eqnarray}
\underset{ \bc}{\min} \: \:  \| \bc^T \bD \|_p^p \quad \text{subject to} \qquad \bc^T \bd_i = 1\:,
\label{eq:main}
\end{eqnarray}
we can conclude that
\begin{eqnarray}
\| \bD^T \bc_i^{*} \|_p \le \| \bD^T \bd_i^{\perp} \|_p \:.
\label{eq:tool}
\end{eqnarray}

\noindent
In addition, 
\begin{eqnarray}
\| \bD^T \bd_i^{\perp} \|_p^p = \| \bD_{k_i}^T \bd_i^{\perp} \|_p^p + \| \bD_{-{k_i}}^T \bd_i^{\perp} \|_p^p \:. 
\label{eq:new_eq1}
\end{eqnarray}
The value of $\| \bD_{k_i}^T \bd_i^{\perp} \|_p^p$ can be upper-bounded as 
\begin{eqnarray}
   \begin{aligned}
\| \bD_{k_i}^T \bd_i^{\perp} \|_p^p & =  \| \bd_i^{\perp} \|_2^p \| \bD_{k_i}^T (\bd_i^{\perp} /  \| \bd_i^{\perp} \|_2)  \|_p^p  \\
 & \le \| \bd_i^{\perp} \|_2^p \underset{\|\bu\| = 1 }{\sup} \:\| \bu^T \dD_{k_i} \|_p^p \\
 & \le \| \bd_i^{\perp} \|_2^p \: \dot{\Delta}_{\max} \:.
     \end{aligned}
     \label{eq:firsttirmnew}
\end{eqnarray}

The columns of $\bD_{-{k_i}}$ lies in a union of $m-1$ linear manifolds and we can rewrite $\| \bD_{-{k_i}}^T \bd_i^{\perp} \|_p^p$ as
\begin{eqnarray}
\| \bD_{-{k_i}}^T \bd_i^{\perp} \|_p^p = \sum_{j \neq k_i} \| \bD_j ^T \bd_i^{\perp} \|_p^p \:,
\end{eqnarray}
using which 
 we can conclude that 
 \begin{eqnarray}
\| \bD_{-{k_i}}^T \bd_i^{\perp} \|_p^p \le (m-1) \max_{j \atop j \neq k_i} (\| \bD_j ^T \bd_i^{\perp} \|_p^p) \:.
\label{eq:ghabli}
\end{eqnarray}
Define $$x = \arg\max_{j \atop j \neq k_i} \: \| \bD_j ^T \bd_i^{\perp} \|_p^p \:. $$
Note that
\begin{eqnarray}
\begin{aligned}
\bD_{x}^T \bd_i^{\perp} & = \left( \bS\bS^T\bD_x + \dU_x\dU_x^T\bD_x \right)^T \bd_i^{\perp}   \\
&= (\dU_x\dU_x^T\bD_x)^T \bd_i^{\perp} \:.
\end{aligned}
\end{eqnarray}
Accordingly, we can rewrite $\| \bD_{x}^T \bd_i^{\perp} \|_p^p$ as 
\begin{eqnarray}
\begin{aligned}
& \| \bD_{x}^T \bd_i^{\perp} \|_p^p = \| (\dU_x\dU_x^T\bD_x)^T \bd_i^{\perp} \|_p^p \\
& =  \| (\dU_x\dU_x^T\bD_x)^T \dU_{k_i} \dU_{k_i}^T \bd_i^{\perp} \|_p^p \\
& \le \| \bd_i^{\perp} \|_2^p \|\dU_{k_i}^T \dU_x \|^p \: \underset{\|\bu\| = 1 \atop \bu \in \mathbb{R}^{r-s}}{\sup} \|  \bu^T \dU_x^T \bD_x \|_p^p \\
 & \le \| \bd_i^{\perp} \|_2^p \max_{j \neq t} \|\dU_t^T \dU_j \|^p
\max_{j} \left( \underset{\|\bu\| = 1}{\sup} \|  \bu^T \dD_j \|_p^p \right) \:.
\label{eq:up-ghoy}
\end{aligned}
\end{eqnarray}
Therefore, according to (\ref{eq:up-ghoy}) and (\ref{eq:ghabli}), 
\begin{eqnarray}
\| \bD_{-{k_i}}^T \bd_i^{\perp} \|_p^p \le (m-1) \| \bd_i^{\perp} \|_2^p  \: \phi^p \: \dot{\Delta}_{\max} \:,
\label{eq:bochek}
\end{eqnarray}
where 
\begin{eqnarray}
\begin{aligned}
& \| \bd_i^{\perp} \|_2^p = \frac{1}{\| \beta_i \|_2^p}\:.
\end{aligned}
\end{eqnarray}

According to (\ref{eq:lhs}), (\ref{eq:firsttirmnew}), and (\ref{eq:bochek}), if
 \begin{eqnarray*}
 \begin{aligned}
&\min_{i} \frac{\| \beta_i \|_2^p}{\| \bd_i \|_2^p} \: \Delta_{\min} \ge \\
& \quad \quad \quad {\dot{\Delta}_{\max} \left( \frac{\kappa}{\kappa + (m-1)} + \kappa \frac{m-1}{\kappa + (m-1)} \phi^p \right) } ,
\end{aligned}
\end{eqnarray*}
then Requirement 1 is satisfied. 
\\
\\
\noindent
\textbf{Proof of Theorem \ref{thm:tsc_deter}}\\
In order to prove that Requirement1 \ref{asm:sufficeint} is satisfied, it is enough to prove that
\begin{eqnarray}
\begin{aligned}
\|  \bd_i^T \bD_{k_i}  \|_p^p \ge \frac{\kappa}{m-1} \| \bd_i^T \bD_{- k_i} \|_p^p \:.
\end{aligned}
\label{eq:newsuff}
\end{eqnarray}

\noindent
In order to ensure that (\ref{eq:newsuff}) holds, we derive the sufficient condition which guarantees that a lower-bound on the LHS of (\ref{eq:newsuff}) is larger than or equal to an upper-bound on the RHS of (\ref{eq:newsuff}). 

First we derive a lower-bound on the LHS. Note that $\bd_i$ and the columns of $\bD_{k_i}$ lie in the same cluster. Accordingly, 
\begin{eqnarray}
\begin{aligned}
\|  \bd_i^T \bD_{k_i}  \|_p^p & \ge \|\bd_i\|_2^p  \underset{\|\bu\| = 1 \atop \bu \in \calS_{k_i}}{\inf} \|  \bu^T \bD_{k_i} \|_p^p \\
& \ge   \|\bd_i\|_2^p  \min_{j} \left\{  \underset{\|\bu\| = 1 \atop \bu \in \calS_{k_j}}{\inf} \|  \bu^T \bD_{k_j} \|_p^p \right\} \:.
\end{aligned}
\label{eq:lowwe}
\end{eqnarray}

Next, we derive an upper-bound on the RHS of (\ref{eq:newsuff}). The matrix $\bD_{- k_i}$  contains the columns of  matrices $\{ \bD_j \}_{j \neq k_i}$. Therefore, 
\begin{eqnarray}
\begin{aligned}
 \| \bd_i^T \bD_{- k_i} \|_p^p & = \sum_{j \neq k_i} \|\bd_i^T \bD_j \|_p^p  \\ &\leq (m-1) \max_{j \atop j \neq k_i} \left\{  \|\bd_i^T \bD_j \|_p^p  \right\} \:.
\end{aligned}
\end{eqnarray}
Define 
\begin{eqnarray}
x = \arg\max_{j \atop j \neq k_i} \: \| \bd_i^T \bD_j  \|_p \:.
\end{eqnarray}
According to the presumed data model, $\bd_i^T \bD_x$ can be expanded as follows
\begin{eqnarray}
\begin{aligned}
\bd_i^T \bD_x & = (\bS \alpha_i + \dU_{k_i} \beta_i)^T (\bS\bS^T 
\bD_x + \dU_x\dU_x^T\bD_x)\\
& =  \alpha_i^T \bS^T \bD_x + \alpha_i^T (\bS^T \dU_x) \dD_x + \beta_i^T (\dU_{k_i}^T \bS) \bS^T \bD_x  \\
& \quad \quad \quad + \beta_i^T (\dU_{k_i}^T \dU_x) \dD_x \\
& = \alpha_i^T \bS^T \bD_x + \beta_i^T (\dU_{k_i}^T \dU_x) \dD_x
\end{aligned}
\label{eq:manyy}
\end{eqnarray}
According to (\ref{eq:manyy}),
 \begin{eqnarray}
\begin{aligned}
& \| \bd_i^T \bD_x \|_p^p  \leq  \|  \alpha_i^T \bS^T \bD_x \|_p^p
+ \| \beta_i^T (\dU_{k_i}^T \dU_x)  \dD_x  \|_p^p  \:.
\end{aligned}
\label{eq:many}
\end{eqnarray}

Next, we establish an upper-bound on each component on the RHS of (\ref{eq:many}). First, we bound $\alpha_i^T \bS^T \bD_x$ as follows
\begin{eqnarray}
\begin{aligned}
& \| \alpha_i^T \bS^T \bD_x\|_p^p = \| \alpha_i^T \bar{\bD}_x\|_p^p = \| \alpha_i \|_2^p \left\| \frac{\alpha_i^T  \bar{\bD}_x}{ \| \alpha_i \|_2}  \right\|_2^p  \\
& \leq \| \alpha_i \|_2^p
\underset{\|\bu\| = 1 \atop \bu \in \mathbb{R}^s}{\sup} \:\| \bu^T \bar{\bD}_x \|_p^p \\
& \leq \| \alpha_i \|_2^p    \:\: \max_{j} \left\{ 
\underset{\|\bu\| = 1 \atop \bu \in \mathbb{R}^s}{\sup} \:\| \bu^T \bar{\bD}_j \|_p^p \right\} \:.
\end{aligned}
\label{eq:firstTerm}
\end{eqnarray}
Finally, the last term of the RHS of (\ref{eq:many}) is upper-bounded as follows,
\begin{eqnarray}
\begin{aligned}
& \| \beta_i^T (\dU_{k_i}^T \dU_x) \dD_x  \|_p^p \leq \\
& \quad \| \beta_i \|_2^p   \max_{j \neq k} \left\{ \| \dU_k^T \dU_j \|^p \right\}    \max_{k} \left\{ 
\underset{\|\bu\| = 1 \atop \bu \in \mathbb{R}^{r-s}}{\sup} \:\| \bu^T {\dD}_k \|_p^p \right\} \:.
\end{aligned}
\label{eq:last_term}
\end{eqnarray}
Therefore, according to (\ref{eq:lowwe}), (\ref{eq:firstTerm}), and (\ref{eq:last_term}), if (\ref{eq:sftc}) holds, then Requirement 1 is satisfied.  
\\
\\
\textbf{Proof of Theorem \ref{thm:random_point}}\\
In order to prove that Requirement 1 is satisfied, we only need to guarantee that (\ref{eq:sfip}) holds. Therefore, a lower bound for $\Delta_{\min}$ and $\min_{i} \frac{\| \beta_i \|_2^p}{\| \bd_i \|_2^p}$ and an upper-bound for $\dot{\Delta}_{\max}$ are established.  
In the presented  proof, we utilize the following lemma whose proof is available in the next section. 

\begin{lemma}
Suppose  $\{\bg_i \in \mathbb{R}^{N}\}_{i=1}^n$   are random  i.i.d. $\calN(0, \frac{1}{N}\bI)$ vectors, i.e., $\mathbb{E}[\bg_i \bg_i^T] = \frac{1}{N} \bI$. If $N > 2$, then
\begin{eqnarray}
\begin{aligned}
& \underset{\|\bu\| = 1}{\sup} \:\: \sum_{i = 1}^{n} (\bu^T \bg_i)^2 \leq \frac{n}{N} + \eta\\
& \underset{\|\bu\| = 1}{\inf} \:\: \sum_{i = 1}^{n} (\bu^T \bg_i)^2 \ge \frac{n}{N} - \eta
\end{aligned}
\label{eq:twokhod}
\end{eqnarray}
with probability at least $1 - 2\delta$ where  
$
 \eta = \max \left( \frac{4 z}{3} \log \frac{2N}{\delta} , \sqrt{4 \frac{n(3 + N)}{N^2} \log \frac{2 N}{\delta}} \right) $, and $
 z = 1 + 2 \sqrt{\frac{1}{N} \log\frac{2n}{\delta} } + \frac{2}{N} \log\frac{2n}{\delta}$.
\label{lm:el2_lm}
\end{lemma}

\noindent
Note that for any $t>0$
\begin{eqnarray}
\begin{aligned}
\mathbb{P} \left[ \Delta_{\min} < t \right] & = \mathbb{P}\left[ \Delta_1 < t \:\: \text{or}\:\: \Delta_2 < t \:\: ... \:\: \text{or} \:\: \Delta_m < t \right] \\
&\le  \sum_{i=1}^m \mathbb{P} \left[ \Delta_i < t\right] \:,
\end{aligned}
\label{eq:ssum_analysis}
\end{eqnarray}
where $\Delta_j =  \underset{\|\bu\| = 1 \atop \bu \in \calS_{j}}{\inf} \:\| \bu^T \bD_{j} \|_p^p$.
According to (\ref{eq:ssum_analysis}), Lemma \ref{lm:el2_lm}, and the definition of ${\eta_{\delta}}_x$,
\begin{eqnarray}
\Delta_{\min}  >  \left(\frac{n}{r} - {\eta_{\delta}}_r \right)
\label{eq:delta_min}
\end{eqnarray}
with probability at least $1 - 2\delta$ because
\begin{eqnarray}
\begin{aligned}
&\mathbb{P} \left[ \Delta_{\min}  < \frac{n}{r} - {\eta_{\delta}}_r \right] \\
&\le \sum_{j} \mathbb{P} \left[ \underset{\|\bu\| = 1 \atop \bu \in \calS_{j}}{\inf} \:\| \bu^T \bD_{j} \|_p^p  < \frac{n}{r} - {\eta_{\delta}}_r \right] \\
& \le \sum_{i=1}^{m }2\delta/m = 2\delta\:.
\end{aligned}
\end{eqnarray}
Similarly, we can conclude that
\begin{eqnarray}
\begin{aligned}
& \dot{\Delta}_{\max}  < \frac{r-s}{r} \Bigg(\frac{n}{r - s} + {\eta_{\delta}}_{r-s}\Bigg)
\end{aligned}
\label{eq:delta_maxdot}
\end{eqnarray}
with probability at least $1 - 2\delta$. 

Next we establish a lower-bound for $\frac{\|\beta_i\|_2^2}{\| \bd_i \|_2^2}$. According to the presumed data model, the distribution of $r \|\beta_i\|_2^2$ is chi-square with $r-s$ degree of freedom and the distribution of $r \| \bd_i \|_2^2$ is chi-square with $r$ degree of freedom.
First, we review the following lemma from \cite{laurent2000adaptive} which provides tail bounds for the chi-square distribution. 
\begin{lemma}
Let $X_r$ be a chi-squared random variable with r-degrees of freedom. Then, for each $t >0$,
\begin{eqnarray}
\begin{aligned}
& \mathbb{P} \left[ X_r  \ge r +  2\sqrt{r t} + 2t \right] \le e^{-t} \:,\\
& \mathbb{P} \left[ X_r \le r - 2\sqrt{r t} \right] \le e^{-t} \:.
\end{aligned}
\end{eqnarray}
\label{lm:lm_xi}
\end{lemma}

\noindent
The distribution of $\frac{\|\beta_i\|_2^2}{\| \bd_i \|_2^2}$ is equivalent to the distribution of $\frac{1}{1 + X_s/X_{r-s}}$ where $X_s$ and $X_{r-s}$ are chi-square random variables with $s$ and $r-s$ degree of freedom, respectively. Therefore,
in order to establish a lower-bound for $\| \beta_i \|_2^2/ \|\bd_i\|_2^2$, we derive an upper-bond for $X_s / X_{r-s}$. 
For any $t_1 > 0$ and $t_2 >0$ we have
\begin{eqnarray}
\begin{aligned}
& \mathbb{P} \left[\frac{X_s}{X_{r-s}} \le \frac{t_1}{t_2} \right] \ge \mathbb{P} \left[X_s \le t_1 \:\: \text{and} \:\: X_{r-s} \ge t_2 \right] \\
& = \mathbb{P} \left[X_s \le t_1 \right] \:  \mathbb{P}\left[X_{r-s} \ge t_2 \right] \:,
\end{aligned}
\label{eq:simple_relue}
\end{eqnarray}
because $X_s$ and $X_{r-s}$ are independent. 
Therefore, according to (\ref{eq:simple_relue}) and according to Lemma  \ref{lm:lm_xi}, if $r - s - 2 \sqrt{(r-s) \log\frac{2M_2}{\delta} }  > 0$, then
\begin{eqnarray}
& \mathbb{P} \left[ \frac{X_{s}}{X_{r-s}} \le \frac{s + 2\sqrt{s \log \frac{2M_2}{\delta}} + 2 \log\frac{2M_2}{\delta}}{ r - s - 2 \sqrt{(r-s) \log\frac{2M_2}{\delta} }} \right] \ge 1 -  \frac{\delta}{M_2} \:.
\label{eq:jadidhateye}
\end{eqnarray}
Using (\ref{eq:jadidhateye}), we can conclude that
\begin{eqnarray}
\begin{aligned}
& \mathbb{P} \Bigg[ \min_{i} \frac{\|\beta_i \|_2^2}{\| \bd_i \|_2^2} \ge \\
& \quad \quad \frac{r - s - 2\sqrt{(r-s) t_{\delta} }}{r + 2\sqrt{s \log \frac{2M_2}{\delta}} + 2 t_{\delta} -  2\sqrt{(r-s) t_{\delta} }} \Bigg]  \ge 1 - \delta \:,
\end{aligned}
\label{eq:beta2}
\end{eqnarray}
where $t_{\delta}=\log\frac{2M_2}{\delta}  $.
Therefore, according to (\ref{eq:delta_min}), (\ref{eq:delta_maxdot}), (\ref{eq:beta2}), and (\ref{eq:sfip}), if the sufficient condition of Theorem \ref{thm:random_point} holds, then Requirement 1 is satisfied with probability at least $1 - 5 \delta$.
\\
\\
\textbf{Proof of Theorem \ref{thm:full_random}}\\
In order to guarantee that Requirement 1 is satisfied with high probability, we only need to guarantee that the sufficient condition of Theorem \ref{thm:random_point} holds. Accordingly, an upper-bound for $\phi$ is derived and it is used to write the new sufficient condition. 
First we review the following lemma whose proof is provided in the next section. 
\begin{lemma}
Suppose that orthonormal matrices $\bU_1$ and $\bU_2$ span two independent randomly generated $r$-dimensional subspaces.  Then,
\begin{eqnarray*}
\mathbb{P} \left[ \| \bU_1^T \bU_2 \| >  \sqrt{\frac{c_{\delta} r^2 }{M_1}}  \right] \le \delta  \:,
\end{eqnarray*}
where $\sqrt{c_{\delta}} = 3 \max \left(1 ,  \sqrt{\frac{8  M_1 \pi }{(M_1 - 1)r}} , \sqrt{\frac{8 M_1 \log r/\delta }{(M_1 - 1)r }} \right) $. 
\label{lm:my_spectral}
\end{lemma}

Let us redefine $c_{\delta} = 3 \max \left(1 ,  \sqrt{\frac{8  M_1 \pi }{(M_1 - 1)(r-s)}} , \sqrt{\frac{16 M_1 \log \frac{m r}{\delta} }{(M_1 - 1)(r-s) }} \right)$.
Therefore,
according to Lemma \ref{lm:my_spectral},
\begin{eqnarray}
\mathbb{P} \left[ \phi >  \sqrt{\frac{c_{\delta} (r-s)^2 }{M_1}}  \right] \le \delta  \:,
\label{eq:upper_phi}
\end{eqnarray}
 because
\begin{eqnarray}
\begin{aligned}
& \mathbb{P} \left[ \phi >  \sqrt{\frac{c_{\delta} (r-s)^2 }{M_1}}  \right] \\
&\le  \sum_{i} \sum_{j \neq i} \mathbb{P} \left[ \|\bU_i \bU_j\| >  \sqrt{\frac{c_{\delta} (r-s)^2 }{M_1}}  \right] \\
& \le \sum_{i} \sum_j \delta/m^2 \:.
\end{aligned}
\label{eq:phi_kol}
\end{eqnarray}
According to (\ref{eq:phi_kol}) and according to Theorem \ref{thm:random_point}, if the sufficient condition of Theorem \ref{thm:full_random} holds, then Requirement 1 is satisfied with probability at least $1- 6\delta$.
\\
\\
\noindent
\textbf{Proof of Theorem \ref{thm:sc_random}}\\
In order to prove that Requirement 1 is satisfied, we only need to guarantee that (\ref{eq:sftc}) holds. Therefore, we establish upper-bounds for 
$\max_{i} \left\{ \frac{\| \alpha_i \|_2^p}{\| \bd_i \|_2^p}  \right\}$, $\max_{i} \left\{ \frac{\| \beta_i \|_2^p}{\| \bd_i \|_2^p}  \right\}$, $\dot{\Delta}_{\max}$, and $\bar{\Delta}_{\max}$ and we establish a lower-bound for ${\Delta}_{\min}$.

As it was shown in the proof of Theorem \ref{thm:random_point},  the distribution of $\| \beta_i\|_2^2/\| \bd_i \|_2^2$ is equivalent to the distribution of $\frac{X_{r-s}}{X_{r-s} + X_{s}}$
where $X_{r-s}$ and $X_{s}$ are chi-squared random variables with $r-s$ and $r$
degrees of freedom, respectively. 
Similarly, the distribution of $\| \alpha_i\|_2^2/\| \bd_i \|_2^2$ is equivalent to the distribution of $\frac{X_{s}}{X_{r-s} + X_{s}}$. 
Therefore, we can use Lemma \ref{lm:lm_xi} to establish an upper-bound for $\max_{i} \left\{ \| \alpha_i \|_2^2 /  \| \bd_i \|_2^2 \right\}$ and an upper-bound for $\max_{i} \left\{ \| \beta_i \|_2^2 / \| \bd_i \|_2^2  \right\}$. 
Note that $\frac{X_{r-s}}{X_{r-s} + X_{s}} = \frac{1}{ 1+ X_{s}/X_{r-s}}$ and $\frac{X_{s}}{X_{r-s} + X_{s}} = \frac{1}{ 1+ X_{r-s}/X_{s}}$ which means that we only need to bound $\frac{X_{s}}{X_{r-s}}$. 
According to Lemma \ref{lm:lm_xi},
\begin{eqnarray*}
\begin{aligned}
& \mathbb{P} \left[ \frac{X_{s}}{X_{r-s}} \le \frac{s - 2\sqrt{s \log \frac{2M_2}{\delta}} }{ r - s + 2 \sqrt{(r-s) \log\frac{2M_2}{\delta} } + 2 \log\frac{2M_2}{\delta}} \right] \\
& \quad \quad \quad \le  \frac{\delta}{M_2} \:, \\
& \mathbb{P} \left[ \frac{X_{r - s}}{X_{s}} \le \frac{(r-s) - 2\sqrt{(r - s) \log \frac{2M_2}{\delta}} }{  s + 2 \sqrt{s \log\frac{2M_2}{\delta} } + 2 \log\frac{2M_2}{\delta}} \right] \le  \frac{\delta}{M_2} \:.
\end{aligned}
\label{eq:two_up}
\end{eqnarray*} 
Therefore, 
\begin{eqnarray}
\begin{aligned}
&\mathbb{P} \Bigg[  \max_{i} \left\{ \| \beta_i \|_2^2 / \| \bd_i \|_2^2  \right\} \ge\\
& \quad \quad \quad \frac{ r - s + 2 \sqrt{(r-s) t_{\delta} } + 2 t_{\delta}}{ r + 2 \sqrt{(r-s) t_{\delta} } + 2 t_{\delta} - 2\sqrt{s\: t_{\delta}}}  
\Bigg] \le \delta \:,
\end{aligned}
\label{eq:11}
\end{eqnarray}
and
\begin{eqnarray}
\begin{aligned}
&\mathbb{P} \Bigg[  \max_{i} \left\{ \| \alpha_i \|_2^2 /  \| \bd_i \|_2^2  \right\} \ge\\
& \quad \quad \quad \frac{ s + 2 \sqrt{s \:t_{\delta} } + 2 t_{\delta} }{r + 2 \sqrt{s \:t_{\delta} } + 2 t_{\delta} - 2\sqrt{(r - s) t_{\delta}}}  
\Bigg] \le \delta \:,
\end{aligned}
\label{eq:12}
\end{eqnarray}
where $t_{\delta} =  \log \frac{2M_2}{\delta}$

According to Lemma \ref{lm:el2_lm} and the definition of $\eta_{\delta_x}$,
\begin{eqnarray}
\bar{\Delta}_{\max}  \le \frac{s}{r} \Bigg(\frac{n}{s} + {\eta_{\delta}}_{s}\Bigg)
\label{eq:13}
\end{eqnarray}
with probability at least $1 - 2\delta$ because
\begin{eqnarray}
\begin{aligned}
&\mathbb{P} \left[  \frac{r}{s} \bar{\Delta}_{\max}  > \frac{n}{s} + {\eta_{\delta}}_{s} \right] \le \\
&\sum_{j} \mathbb{P} \Bigg[ \frac{r}{s}\: \underset{\|\bu\| = 1}{\sup} \:\| \bu^T \bar{\bD}_{j} \|_p^p  > \frac{n}{s} + {\eta_{\delta}}_{s} \Bigg] \le \sum_{j=1}^m \frac{2\delta}{m}\:.
\end{aligned}
\end{eqnarray}
Similarly, 
\begin{eqnarray}
\begin{aligned}
& \mathbb{P} \Bigg[\dot{\Delta}_{\max}  > \frac{r-s}{r} \left( \frac{n}{r-s} +{\eta_{\delta}}_{r-s}\right)\Bigg] < 2\delta \:.
\end{aligned}
\label{eq:14}
\end{eqnarray}
Therefore, according to (\ref{eq:delta_min}), (\ref{eq:phi_kol}), (\ref{eq:11}), (\ref{eq:12}), (\ref{eq:13}), and (\ref{eq:14}), if the sufficient condition of Theorem \ref{thm:sc_random} holds, 
Requirement 1 is satisfied with probability at $1 - 9 \delta$.
\\
\\
\textbf{Proof of Theorem \ref{thm:with_inv}}\\
We use the same procedure employed in the proof of Theorem \ref{thm: 1}, i.e.,  a sufficient condition is derived which guarantees that 
\begin{eqnarray}
\| \bD_{k_i}^T \bc_i^{*} \|_p^p > \frac{\kappa}{\kappa + m-1} \| \bD^T \bc_i^{*} \|_p^p 
\label{eq:anotherone}
\end{eqnarray} 
holds for all $ 1\le i \le M_2$.
The lower-bound on the LHS of  (\ref{eq:anotherone}) was derived in (\ref{eq:lhs}). 

We only need to establish an upper-bound for the RHS of  (\ref{eq:anotherone}). Corresponding to each data point $\bd_i$, define $\Vec{\bd}_i$ as 
\begin{eqnarray}
\Vec{\bd}_i = \frac{\vec{\bU}_{k_i}\vec{\bU}_{k_i}^T \bd_i}{\| \vec{\bU}_{k_i}^T \dot{\bU}_{k_{i}} \beta_i \|_2^2} \:.
\end{eqnarray}
Note that $\vec{\bd}_i^T \bd_i =1$ because $\dU_{k_i}$ and $\vec{\bU}_{k_i}$ are orthogonal to $\calS$ and
\begin{eqnarray}
\vec{\bd}_i^T \bd_i = \frac{\bd_i^T \vec{\bU}_{k_i}\vec{\bU}_{k_i}^T \bd_i}{\| \vec{\bU}_{k_i}^T \dot{\bU}_{k_{i}} \beta_i \|_2^2} = \frac{\beta_i^T \dU_{k_i}^T \vec{\bU}_{k_i}\vec{\bU}_{k_i}^T \dU_{k_i} \beta_i}{\| \vec{\bU}_{k_i}^T \dot{\bU}_{k_{i}} \beta_i \|_2^2} = 1 \:.
\end{eqnarray}
Since $\vec{\bd}_i^T \bd_i =1$ and $\bc_i^{*}$ is the optimal direction, we conclude that
\begin{eqnarray}
\| \bD^T \bc_i^{*} \|_p^p  \le \| \bD_{k_i}^T \vec{\bd}_i \|_p^p + \| \bD_{-{k_i}}^T \vec{\bd}_i \|_p^p \:.
\end{eqnarray}
In addition, $\vec{\bd}_i$ is orthogonal to $\oplus_{j\neq k_i} \calS_i$ which means that $\| \bD_{-{k_i}}^T \vec{\bd}_i \|_p^p = 0$. 
Moreover, since $\vec{\bd}_i$ is orthogonal $\calS$,
\begin{eqnarray}
\begin{aligned}
   \| \bD_{k_i}^T \vec{\bd}_i \|_p^p  & = \| \dD_{k_i}^T \dU_{k_i} \dU_{k_i}^T \vec{\bd}_i \|_p^p \\
 & \le \| \vec{\bd}_i\|_2^p  \underset{\|\bu\| = 1}{\sup} \:\| \bu^T \dD_{k_i} \|_p^p\:.
  \end{aligned}
\end{eqnarray}
In addition,
\begin{eqnarray}
\begin{aligned}
\| \vec{\bd}_i\|_2 = \frac{1}{\| \vec{\bU}_{k_i}^T \dot{\bU}_{k_{i}} \beta_i \|_2} \:.
  \end{aligned}
\end{eqnarray}
Therefore, if 
\begin{eqnarray}
\begin{aligned}
 & \frac{1}{\| \bd_i \|_2^p}\underset{\|\bu\| = 1 \atop \bu \in \calS_{k_i}}{\inf} \|\bu^T \bD_{k_i} \|_p^{p} \: \| \vec{\bU}_{k_i}^T \dot{\bU}_{k_{i}} \beta_i \|_2^p \ge \\
 & \quad \quad \frac{\kappa}{\kappa + m-1} \:  \underset{\|\bu\| = 1}{\sup} \:\| \bu^T \dD_{k_i} \|_p^p \:,
  \end{aligned}
  \label{eq:inv_suff}
\end{eqnarray}
then (\ref{eq:anotherone}) holds. 
In addition,
\begin{eqnarray}
\begin{aligned}
 \| \vec{\bU}_{k_i}^T \dot{\bU}_{k_{i}} \beta_i \|_2 \ge  \| \beta_i \|_2  \| \vec{\bU}_{k_i}^T \dot{\bU}_{k_{i}} \|_{m} \:,
  \end{aligned}
\end{eqnarray}
where $\| \vec{\bU}_{k_i}^T \dot{\bU}_{k_{i}} \|_{m}$ denotes the minimum singular value of matrix $\vec{\bU}_{k_i}^T \dot{\bU}_{k_{i}}$. 
Accordingly, if (\ref{eq:withinnovsuff}) holds, 
then Requirement 1 is satisfied. 
\\
\\
\textbf{Proof of Theorem \ref{thm:with_inv_random}}\\
According to the proof of Theorem \ref{thm:with_inv}, it is enough to guarantee that (\ref{eq:inv_suff}) holds for all the data points. 
First we establish a lower-bound for $$\| \vec{\bU}_{k_i} \vec{\bU}_{k_i}^T \dot{\bU}_{k_{i}} \beta_i \|_2 / \|\bd_i\|_2^2 \:.$$
Note that $\vec{\bU}$ is the span of $\{ (\bI - \bP_{k_i} \bP_{k_i}^T) \bx \:\:| \:\: \bx \in \dot{\calS}_{k_i} \}$ where $\bP_{k_i}$ is an orthonormal basis for $\oplus_{j \neq k_i} \calS_{j}$. Thus, the column space of $\vec{\bU}_{k_i}$
is a subset of the column space of $(\bI - \bP_{k_i} \bP_{k_i}^T)$. Therefore,
$\vec{\bU}_{k_i} \vec{\bU}_{k_i}^T \dot{\bU}_{k_{i}} \beta_i  = (\bI - \bP_{k_i} \bP_{k_i}^T) \dot{\bU}_{k_{i}} \beta_i $. Since the subspaces $\{ \dot{\calS}_{i} \}_{i=1}^m$ and $\calS$ are generated independently and  uniformly at random
and $M_1 > s + (r-s)m$, then $\{ \dot{\calS}_{i} \}_{i=1}^m$ and $\calS$ are
independent subspaces with an overwhelming probability. In other word, the dimension of  
$(\oplus_{i=1}^{m}\dot{\calS}_{i=1}) \oplus \calS$ is equal to $s + (r-s)m$  with an overwhelming probability \cite{vershynin2010introduction}. Let us assume that this is true using which we can conclude that 
 the column-space of $(\bI - \bP_{k_i} \bP_{k_i}^T)$ is a random $s + (r-s)(m-1)$ subspace which is generated independently from $\dot{\calS}_{k_i}$.
 Define $\bH_{k_i}$ as an orthonormal basis for $(\bI - \bP_{k_i} \bP_{k_i}^T)$.
Since the column space of $\bH_{k_i}$ and $\dot{\calS}_{k_i}$  
are generated independently, the distribution of $$\frac{\| \bH_{k_i}^T \dU_{k_i} \beta_i\|_2^2}{\|\beta_i \|_2^2 + \|\alpha_i \|_2^2}$$ is equivalent to the distribution of 
$$\frac{\| \bH_{k_i}^T \ba \|_2^2 \: \| \beta_i\|_2^2}{\|\beta_i \|_2^2 + \|\alpha_i \|_2^2}$$
where $\ba \in \mathbb{R}^{M_1}$ is a random vector on $\mathbb{S}^{M_1 - 1}$ which is independent from $\| \beta_i \|_2^2$. In addition, for any $t_1 > 0$ and $t_2 >0$ we can write
\begin{eqnarray}
\begin{aligned}
& \mathbb{P} \left[ \frac{\| \bH_{k_i}^T \ba \|_2^2 \: \| \beta_i\|_2^2}{\|\beta_i \|_2^2 + \|\alpha_i \|_2^2} \ge t_1 \: t_2 \right] \ge\\
& \quad \quad \mathbb{P} \left[ \frac{ \| \beta_i\|_2^2}{\|\beta_i \|_2^2 + \|\alpha_i \|_2^2} \ge t_1 \quad \text{and} \quad \| \bH_{k_i}^T \ba \|_2^2 \ge t_2\right] \:.
  \end{aligned}
 \label{eq:dota1}
\end{eqnarray}
Similar to (\ref{eq:11}), $$\mathbb{P} \left[ \frac{ \| \beta_i\|_2^2}{\|\beta_i \|_2^2 + \|\alpha_i \|_2^2} \ge t_1 \right] \ge 1 - \delta/M_2$$ where $t_1 = \frac{r - s - 2\sqrt{(r-s) \log\frac{2M_2}{\delta} }}{r + 2\sqrt{s \log \frac{2M_2}{\delta}} + 2 \log\frac{2M_2}{\delta} -  2\sqrt{(r-s) \log\frac{2M_2}{\delta} }} $. 

Since the column-space of $\bH_{k_i}$ is a random subspace and $\ba$ is a random vector on $\mathbb{S}^{M_1 -1}$, the distribution of $\| \bH_{k_i}^T \ba \|_2$ is equivalent to the distribution of $\| \bH_{k_i}^T \be_1 \|_2$ where $\be_1$ is the first column of the identity matrix $\bI \in \mathbb{R}^{M_1 \times M_1}$. The distribution of $\| \bH_{k_i}^T \be_1 \|_2^2$
is equivalent to the distribution of $\frac{X_{M_1 - \vartheta }}{X_{M_1}}$ where $X_{M_1}$ is a chi-squared  random  variables  with $M_1$ degree of freedom and $\vartheta	 = M_1 - \big(s + (r-s)(m-1)\big)$ \cite{candes2009exact}. Therefore,  we can bound $\| \bH_{k_i}^T \be_1 \|_2^2$ similarly
\begin{eqnarray}
\begin{aligned}
& \mathbb{P} \Bigg[ \| \bH_{k_i}^T \be_1 \|_2^2 \ge t_2 \Bigg]  \ge 1 - \frac{\delta}{M_2} \:,
\end{aligned}
\label{eq:dota2}
\end{eqnarray}
where $t_2 = \frac{\vartheta - 2\sqrt{\vartheta\: t_{\delta} }}{M_1 + 2\sqrt{(M_1 - \vartheta) t_{\delta} } + 2t_{\delta}  -  2\sqrt{\vartheta t_{\delta}  }}$ and $t_{\delta} = \log \frac{2 M_2}{\delta}$.
Therefore, according to (\ref{eq:delta_min}), (\ref{eq:delta_maxdot}),  (\ref{eq:inv_suff}), (\ref{eq:dota1}), and (\ref{eq:dota2}), if the sufficient condition of Theorem \ref{thm:with_inv_random} holds, then Requirement 1 is satisfied with probability at least $1 - 6 \delta - \epsilon$ where $\epsilon$ is the probability that the rank of $\bD$ is smaller than $s + (r-s)m$.

\section{Proofs of the Intermediate Results}
\noindent
\textbf{Proof of Lemma \ref{lm:el2_lm} }\\
First, we add and subtract the mean of each random component as follows
\begin{eqnarray}
\begin{aligned}
&\underset{\|\bu\| = 1}{\sup} \:\: \sum_{i = 1}^{n} (\bu^T \bg_i)^2 \leq \\
& \quad \underset{\|\bu\| = 1}{\sup} \:\: \sum_{i = 1}^{n} \left[ (\bu^T \bg_i)^2 -  \mathbb{E} (\bu^T \bg_i)^2  \right]  + \underset{\|\bu\| = 1}{\sup} \:\: \sum_{i = 1}^{n} \mathbb{E} (\bu^T \bg_i)^2.
\label{eq:main_1nq}
\end{aligned}
\end{eqnarray}
Note that 
\begin{eqnarray}
\begin{aligned}
\mathbb{E} (\bu^T \bg_i)^2 & = \frac{1}{N}\mathbb{E} \: r_i (\bu^T \bs_i)^2  = \frac{1}{N} \mathbb{E} \: r_i (\bu^T \bs_i )^2  \\
&= \frac{1}{N} \mathbb{E} \: r_i \mathbb{E} (\bu^T \bs_i )^2
\end{aligned}
\end{eqnarray}
where $\bs_i \in \mathbb{R}^{N}$ is a random vector on $\mathbb{S}^{N -1}$ and $r_i$ is a chi-squared  random variable with $N$ degree of freedom. Since $\bs_i$ is sampled uniformly at random from $\mathbb{S}^{N -1}$,
\begin{eqnarray}
\mathbb{E} (\bu^T \bs_i )^2 = \mathbb{E} (\be_1^T \bs_i )^2\:,
\end{eqnarray}
where $\be_i$ is the first column of  Identity matrix $\bI$. Therefore,
\begin{eqnarray}
\begin{aligned}
& \mathbb{E} (\bu^T \bg_i)^2 = \frac{1}{N}\mathbb{E} \: r_i \:\: \mathbb{E} (\be_1^T \bs_i )^2 = \mathbb{E} (\be_1^T \bg_i )^2 = \frac{1}{N} \:,
\end{aligned}
\end{eqnarray}
and
\begin{eqnarray}
\underset{\|\bu\| = 1}{\sup} \:\: \sum_{i = 1}^{n} \mathbb{E} (\bu^T \bg_i)^2 = \frac{n}{N} \:.
\label{mean_mylemaj}
\end{eqnarray}
The first component of the RHS of (\ref{eq:main_1nq}) can be rewritten as
\begin{eqnarray}
\begin{aligned}
 & \underset{\|\bu\| = 1}{\sup} \:\: \sum_{i = 1}^{n} \left[ (\bu^T \bg_i)^2 -  \mathbb{E} (\bu^T \bg_i)^2  \right]  \\
 & = \underset{\|\bu\| = 1}{\sup} \:\:
 \bu^T  \left( \sum_{i = 1}^{n} \bg_i \bg_i^T - \mathbb{E} \: \{ \bg_i \bg_i^T \}     \right) \bu \\
 & = \underset{\|\bu\| = 1}{\sup} \:\:
 \bu^T  \left( \sum_{i = 1}^{n} \bg_i \bg_i^T -  \frac{1}{N}\bI     \right) \bu.
\end{aligned}
\end{eqnarray}
The matrices $\{ \bg_i \bg_i^T - \frac{1}{N}\bI \}_{i = 1}^n$ are zero mean random matrices. Thus, we use the non-commutative Bernstein inequality to bound the spectral norm of the matrix  $\bM$ defined as
\begin{eqnarray}
\bM =  \sum_{i = 1}^{n} \left( \bg_i \bg_i^T - \frac{1}{N} \bI    \right).
\end{eqnarray}

\begin{lemma} \cite{junge2013noncommutative}
Let $\bX_1 , \bX_2 , ... , \bX_L $ be independent zero-mean random matrices of dimension $d_1 \times d_2$. Suppose $\rho_k^2 = \max \{\| \mathbb{E} [ \bX_k \bX_k^T ] \| , \| \mathbb{E} [ \bX_k^T \bX_k ] \| \} $ and $\| \bX_k \| \leq M$ almost surely for all k. Then for any $\tau > 0$
\begin{eqnarray}
\begin{aligned}
&\mathbb{P} \left[ \Bigg \| \sum_{k=1}^{L} \bX_k \Bigg  \| > \tau \right]   \leq  \\
& \quad \quad \quad (d_1 + d_2) \exp \left( \frac{-\tau^2 / 2}{\sum_{k=1}^L \rho_k^2  + M\tau/3} \right) .
\end{aligned}
\label{eq28}
\end{eqnarray}
\label{lm5}
\end{lemma}

\noindent
To find the parameter $M$ defined in Lemma \ref{lm5}, we compute
\begin{eqnarray}
\begin{aligned}
 & \| \bg_i \bg_i^T -  \frac{1}{N}\bI \| \leq \max ( \| \bg_i \bg_i^T \| ,\frac{1}{N})
\end{aligned}
\end{eqnarray}
where we used the fact that $ \| \bH_1 - \bH_2  \| \leq \max ( \|\bH_1 \| , \|\bH_2 \|)$, if $\bH_1$ and $\bH_2$ are positive definite matrices.
Note that,
\begin{eqnarray}
\begin{aligned}
\| \bg_i \bg_i^T \| = \underset{\|\bu\| = 1}{\sup} \| \bg_i \bg_i^T \bu\|_2 \le \| \bg_i \|_2^2 \:.
\end{aligned}
\end{eqnarray}
The distribution of $N \| \bg_i \|_2^2$ is equivalent to the distribution of a chi-square distribution with $N$ degrees of freedom. Therefore, according to Lemma \ref{lm:lm_xi},
\begin{eqnarray}
\begin{aligned}
\mathbb{P} \left[ \| \bg_i \bg_i^T \| >  1 + 2 \sqrt{\frac{1}{N} \log\frac{2n}{\delta} } + \frac{2}{N} \log\frac{2n}{\delta} \right] < \frac{\delta}{n} \:.
\end{aligned}
\label{eq:zetaha}
\end{eqnarray}
According to (\ref{eq:zetaha}) we conclude that $M$ is less than or equal to $1 + 2 \sqrt{\frac{1}{N} \log\frac{2n}{\delta} } + \frac{2}{N} \log\frac{2n}{\delta}$ with probability at least $1 - \delta$.  

For the parameter $\rho$ we have
\begin{eqnarray}
\begin{aligned}
 & \left\| \mathbb{E}  \left[  \left( \bg_i \bg_i^T -  \frac{1}{N} \bI \right) \left( \bg_i \bg_i^T - \frac{1}{N}\bI \right) \right] \right\| = \\
 &  \left\| \mathbb{E}  \left[ \bg_i \bg_i^T \bg_i \bg_i^T - \frac{2}{N} \bg_i \bg_i^T  +  \frac{1}{N^2}\bI \right] \right\| = \\
 & \left\| \mathbb{E} \: \left[ \bg_i \bg_i^T \bg_i  \bg_i^T\right] - \frac{2}{N} \bg_i \bg_i^T  +  \frac{1}{N^2}\bI  \right\| = \\
 & \left\| \mathbb{E} \left[ \bg_i \bg_i^T \bg_i \bg_i^T \right] -  \frac{1}{N^2}\bI  \right\| \le \max\left(\frac{1}{N^2},  \left\| \mathbb{E}  \left[ \bg_i \bg_i^T \bg_i \bg_i^T \right] \right\| \right)
\end{aligned}
\label{eq:chandjomle}
\end{eqnarray}
Therefore, we only need to compute $\left\| \mathbb{E}  \left[ \bg_i \bg_i^T \bg_i \bg_i^T \right] \right\| $. Define $\bG = N^2 \bg_i \bg_i^T \bg_i \bg_i^T$. The distribution of diagonal values of $\bG$ is equivalent to the distribution of $X_1^2$ where $X_1$ a chi squared random variable with 1 degree of freedom. Thus, $\mathbb{E} \left[ \bG(i,i) \right] = 3$. The distribution of  off-diagonal elements of $\bG$ is equivalent to $X_1 Y_1$ where $X_1$ and $Y_1$ are independent chi squared random variables with 1 degree of freedom. Thus, $\mathbb{E} \left[ \bG(i,i) \right] = 1$ when $i \neq j$. Accordingly, 
\begin{eqnarray}
N^2 \mathbb{E}  \left[ \bg_i \bg_i^T \bg_i \bg_i^T \right] = 2 \bI + \mathbf{1} \mathbf{1}^T \:,
\end{eqnarray}
where $\mathbf{1} \in \mathbb{R}^{N}$ is a vector whose all elements are equal to 1. Therefore, \textcolor{black}{according to $\| \mathbb{E}  \left[ \bg_i \bg_i^T \bg_i \bg_i^T \right] \| = \frac{2 + N}{N^2}$ and (\ref{eq:chandjomle})}, $$\left\| \mathbb{E}  \left[  \left( \bg_i \bg_i^T -  \frac{1}{N} \bI \right) \left( \bg_i \bg_i^T - \frac{1}{N}\bI \right) \right] \right\| \le \frac{3 + N}{N^2} \:.$$

\noindent
According to Lemma \ref{lm5} and assuming that $\| \bg_i \bg_i^T -  \frac{1}{N}\bI \| \le z_{\delta}$,
\begin{eqnarray}
\begin{aligned}
&\mathbb{P} \left[ \| \bM \| > \tau \right] \leq
 2 N \exp \left( \frac{-\tau^2 / 2}{n \frac{3 + N}{N^2}  +  z_{\delta} \tau/3} \right) \:.
\end{aligned}
\end{eqnarray}
In addition, 
\begin{eqnarray}
\begin{aligned}
\exp \left( \frac{-\tau^2 / 2}{n \frac{3 + N}{N^2}  +  z_{\delta} \tau/3} \right) \le  \left( \frac{-\tau^2 / 2}{2\max(n \frac{3 + N}{N^2}  ,  z_{\delta} \tau/3)} \right)
\end{aligned}
\end{eqnarray}
Thus,
\begin{eqnarray}
\begin{aligned}
&\mathbb{P} \left[ \| \bM \| > \eta_{\delta} \right] \leq
2 \delta
\end{aligned}
\label{eq:uperbound_mylm}
\end{eqnarray}
where
\begin{eqnarray}
\begin{aligned}
& \eta_{\delta} = \max \left( \frac{4 z_{\delta}}{3} \log \frac{2N}{\delta} , \sqrt{4 \frac{n(3 + N)}{N^2} \log \frac{2 N}{\delta}} \right) \\
& z_{\delta} = 1 + 2 \sqrt{\frac{1}{N} \log\frac{2n}{\delta} } + \frac{2}{N} \log\frac{2n}{\delta} \:.
\end{aligned}
\label{eta_def}
\end{eqnarray}
According to (\ref{mean_mylemaj}) and (\ref{eq:uperbound_mylm}),
\begin{eqnarray}
\begin{aligned}
& \underset{\|\bu\| = 1}{\sup} \:\: \sum_{i = 1}^{n} (\bu^T \bg_i)^2 <  \frac{n}{N} + \eta_{\delta}
\end{aligned}
\end{eqnarray}
with probability at least $1 - 2\delta$.

We use similar techniques to prove the second inequality in (\ref{eq:twokhod}). First, we add and subtract the mean of each random component as follows
\begin{eqnarray}
\begin{aligned}
& \underset{\|\bu\| = 1}{\inf} \:\: \sum_{i = 1}^{n} (\bu^T \bg_i)^2\ge  \underset{\|\bu\| = 1}{\inf} \:\: \sum_{i = 1}^{n} \mathbb{E} (\bu^T \bg_i)^2 +\\
& \quad \quad \quad \underset{\|\bu\| = 1}{\inf} \:\: \sum_{i = 1}^{n} \left[ (\bu^T \bg_i)^2 -  \mathbb{E} (\bu^T \bg_i)^2  \right] \:.
\label{eq:main_1nq}
\end{aligned}
\end{eqnarray}
In addition, $\underset{\|\bu\| = 1}{\inf} \:\: \sum_{i = 1}^{n} \left[ (\bu^T \bg_i)^2 -  \mathbb{E} (\bu^T \bg_i)^2  \right] = -1 \times \underset{\|\bu\| = 1}{\sup} \:\: \sum_{i = 1}^{n} \left[ \mathbb{E} (\bu^T \bg_i)^2 -  (\bu^T \bg_i)^2  \right]$ which can be  rewritten as 
\begin{eqnarray}
\begin{aligned}
& \underset{\|\bu\| = 1}{\sup} \:\: \sum_{i = 1}^{n} \left[ \mathbb{E} (\bu^T \bg_i)^2 -  (\bu^T \bg_i)^2  \right] = \\
&\quad \quad \quad \left \| \sum_{i = 1}^{n} \left( \bg_i \bg_i^T -  \frac{1}{N} \bI    \right)  \right\| \:.
\end{aligned}
\end{eqnarray}
Using similar techniques which was used to bound $\|\bM \|$, we can conclude that
\begin{eqnarray}
\begin{aligned}
\underset{\|\bu\| = 1}{\sup} \:\: \sum_{i = 1}^{n} \left[ \mathbb{E} (\bu^T \bg_i)^2 -  (\bu^T \bg_i)^2  \right] \le \eta_{\delta}
\end{aligned}
\label{eq:mines_1}
\end{eqnarray}
with probability at least $1 - 2\delta$. Thus, according to (\ref{eq:main_1nq}) and (\ref{eq:mines_1}),
\begin{eqnarray}
\begin{aligned}
& \underset{\|\bu\| = 1}{\inf} \:\: \sum_{i = 1}^{n} (\bu^T \bg_i)^2 >  \frac{n}{N} - \eta_{\delta}
\end{aligned}
\end{eqnarray}
with probability at least $1 - 2\delta$.
\\
\\
\textbf{Proof of Lemma \ref{lm:my_spectral}}\\
First we review the following lemma.
\begin{lemma}
\cite{park2014greedy,ledoux2005concentration}
Let the columns of $\bU \in \mathbb{R}^{N \times {r}}$ be an orthonormal basis for an ${r}$-dimensional random subspace drawn uniformly at random in an ambient $N$-dimensional space. For a unit $\ell_2$-norm vector $\bc \in \mathbb{R}^{N \times 1}$
\begin{eqnarray*}
\mathbb{P} \left[ \| \bc^T \bU \|_2 > \sqrt{\frac{{ c_{\delta} r}}{N}}  \right] \leq \delta \:,
\end{eqnarray*}
where $\sqrt{c_{\delta}} = 3 \max \left(1 ,  \sqrt{\frac{8  N \pi }{(N - 1)r}} , \sqrt{\frac{8 N \log1/\delta }{(N - 1)r}} \right)$.
\label{lm:projectranodm}
\end{lemma}

The spectral norm of matrix $ \bU_1^T \bU_2$ is equal to its largest singular value. 
Define  $\bs \in \mathbb{R}^r$ as the vector of singular values of $\bU_1^T \bU_2$ where $\max_{i} \bs(i) = \bs(1)$. 
Note that
\begin{eqnarray}
\begin{aligned}
& \| \bU_1^T \bU_2 \| = \bs(1) \le \| \bs \|_2 = \|\bU_1^T \bU_2 \|_F
\end{aligned}
\end{eqnarray}
where $\| \bU_1^T \bU_2 \|_F$ is the Frobenius norm of $\bU_1^T \bU_2$. 
The Frobenius norm can be expanded as
\begin{eqnarray}
\begin{aligned}
\| \bU_1^T \bU_2  \|_F^2 = \sum_{i=1}^{r} \| {\mathbf{u}_1}_i^T \bU_2  \|_2^2 \:,
\end{aligned}
\end{eqnarray}
where 
${\mathbf{u}_1}_i$ is the $i^{th}$ column of $\bU_1$. According to Lemma \ref{lm:projectranodm}, 
\begin{eqnarray}
\begin{aligned}
\mathbb{P} \left[ \| {\mathbf{u}_1}_i^T \bU_2  \|_2^2 >  \frac{r \: c_{\frac{\delta}{r}} }{N}   \right] \le \frac{\delta}{r} \:.
\end{aligned}
\end{eqnarray}
Therefore, 
\begin{eqnarray}
\begin{aligned}
\| \bU_1^T \bU_2  \|_F^2  \le \frac{c_{\frac{\delta}{r}} \:r^2 \:}{N} 
\end{aligned}
\end{eqnarray}
with probability at least $1 - \delta$.

\end{document}